\documentclass{mfc}
\usepackage{amsmath}
\usepackage{amssymb}
\usepackage{comment}
\usepackage{epsfig}
\usepackage{epstopdf}
\usepackage{graphics}
\usepackage{graphicx}
\usepackage{paralist}
\usepackage{soul}
\usepackage{xcolor}
\usepackage[colorlinks=true]{hyperref}
\DeclareMathOperator*{\argmax}{argmax}
\DeclareMathOperator*{\argmin}{argmin}

\hypersetup{urlcolor=blue, citecolor=red}
\def\currentvolume{1}
 \def\currentissue{2}
  \def\currentyear{2018}
   \def\currentmonth{May}
    \def\ppages{149--180}
     \def\DOI{10.3934/mfc.2018008}

\theoremstyle{definition}

\title[How Convolutional Neural Networks See the World]
	{How convolutional neural networks see the world --- A survey of convolutional neural network visualization methods}

\author[Zhuwei Qin, Funxun Yu, Chenchen Liu and Xiang Chen]{}


\email{zqin@gmu.edu}
\email{fyu2@gmu.edu}
\email{chliu@clarkson.edu}
\email{xchen26@gmu.edu}

\thanks{The authors are supported by NSF Grant CNS-1717775}

\thanks{$^*$ Corresponding Author: Xiang Chen}

\begin{document}
\maketitle

\centerline{\scshape Zhuwei Qin$\dagger$}
\medskip
{\footnotesize
 \centerline{$\dagger$George Mason University}
   \centerline{4400 University Dr, Fairfax, VA 22030, USA}
}
\medskip

\centerline{\scshape Fuxun Yu$\dagger$, Chenchen Liu$\ddagger$ and Xiang Chen$\dagger$$^*$}
\medskip
{\footnotesize
\centerline{$\ddagger$Clarkson University}
   \centerline{8 Clarkson Ave, Potsdam, NY 13699, USA}
}
\bigskip

 \centerline{(Communicated by Zhipeng Cai)}

\begin{abstract}
Nowadays, the Convolutional Neural Networks (CNNs) have\break achieved impressive performance on many computer vision related tasks, such as object detection, image recognition, image retrieval, \textit{etc}.
	These achievements benefit from the CNNs' outstanding capability to learn the input features with deep layers of neuron structures and iterative training process.
	However, these learned features are hard to identify and interpret from a human vision perspective, causing a lack of understanding of the CNNs' internal working mechanism.
To improve the CNN interpretability, the CNN visualization is well utilized as a qualitative analysis method, which translates the internal features into visually perceptible patterns.
	And many CNN visualization works have been proposed in the literature to interpret the CNN in perspectives of network structure, operation, and semantic concept.

In this paper, we expect to provide a comprehensive survey of several representative CNN visualization methods, including \textit{Activation Maximization}, \textit{Network Inversion}, \textit{Deconvolutional Neural Networks} (\textit{DeconvNet}), and \textit{Network Dissection} based visualization.
	These methods are presented in terms of motivations, algorithms, and experiment results.
	Based on these visualization methods, we also discuss their practical applications
	to demonstrate the significance of the CNN interpretability in areas of network design, optimization, security enhancement, \textit{etc}.
\end {abstract}
\section{Introduction}
The Convolutional Neural Networks (CNNs) have been widely investigated as one of the most promising solutions for various computer vision related tasks, such as object detection~\cite{Girsh:2014:CVPR,Ren:2015:NIPS}, image recognition~\cite{Kriz:2012:NIPS:Alexnet,Simonyan:2014:VGG,Szegedy:2015:CVPR:GoogleNet,Kai:2016:ResNet}, image retrieval~\cite{Gong:2014:ECCV,Gordo:2016:deep-retrieval}, \textit{etc}.

Inspired by the hierarchical organization of the human visual cortex~\cite{cortex}, the CNN is constructed with many intricately interconnected layers of neuron structures. These neurons act as the basic units to learn and extract certain features from the input.
	With the network complexity increment caused by the neuron layer depth,  the performance of the input feature extraction is also enhanced. For example, \textit{AlexNet} -- one of the most representative CNNs, has 650K neurons and 60M related parameters~\cite{Kriz:2012:NIPS:Alexnet}.
	Also, sophisticated algorithms are proposed to support the training and testing of such a complex network, and the backpropagation method is widely applied to train the CNN parameters through multiple layers~\cite{Lecun:1998:backpropgation,Lin:microsoftCOCO}.
	Furthermore, to fine-tune the network to specific functions, large pools of labeled data are required for iteratively training the massive neurons and connection weights.
	So far, many high-performance CNN designs have been proposed, such as \textit{AlexNet}~\cite{Kriz:2012:NIPS:Alexnet}, \textit{VGG}~\cite{Simonyan:2014:VGG}, \textit{GoogleNet}~\cite{Szegedy:2015:CVPR:GoogleNet}, \textit{ResNet}~\cite{Kai:2016:ResNet}, \textit{etc}. Some of the designs can even achieve beyond human-level accuracy on object recognition tasks~\cite{Silver:2016:alphago}.

Although the CNNs can achieve competitive classification accuracy, the CNNs still suffer from high computational cost, slow training speed, and security vulnerability~\cite{Szegedy:2013:adversarial,Kurakin:2016:adversarial,Ciresan:2011:flexible}.
	One major reason causing these shortcomings is the lack of network interpretability, especially the limited understanding of the internal features learned by each convolutional layer:
	Mathematically, the convolutional layer neurons (namely the convolution filters) convolve with the input image or the outputs of the previous layer, the results are considered as learned features and recorded in the feature maps. With deeper layers, the neurons are expected to extract higher level features, and eventually converge to the final classification.
	However, as the CNN training is considered as a black-box process and the neurons are designed in the format of simple matrices, the formation of those neuron values are unpredictable and the neuron meanings are impossible to directly explained.
	Hence, the poor network interpretability significantly hinders the robustness evaluation of each network layer, the further optimization on the network structure, as well as the network adaptability and transferability to different applications~\cite{Pan:2010:transfer-learning,Shin:2016:transfer-learning}.

A qualitative way to improve the network interpretability is the network visualization, which translates the internal features into visually perceptible image patterns.
	This visualization process is referred from the human visual cortex system analysis:
	In a human brain, the human visual cortex is embedded in multiple vision neuron areas~\cite{detector}.
	In each vision neuron area, numerous neurons selectively respond to different features, such as colors, edges, and shapes~\cite{Kay:2008:identifying,Posner:1990:attention}.
	To explore the relationship between the neurons and features, researchers usually find the preferred stimulus to identify individual kind of the response and illustrate the response to certain visual patterns.
	The CNN visualization also follows such an analytical approach to realize the CNN interpretability.

Up to now, many effective network visualization works have been proposed in the literature, and several representative methods are widely adopted:
	1) Erhan \textit{et al.} proposed the \textit{Activation Maximization} to interpret traditional shallow networks~\cite{Erhan:2009:ICML:AM,Hinton:2006:DBNs,Vincent:2010:SDAE}.
	Later, this method was further improved by Simonyan \textit{et al.}, which synthesized an input image pattern with the maximum activation of a single CNN neuron for visualization~\cite{Simonyan:2013:ICLR:AM}.
	This fundamental method was also extended by many other works with different regularizers for interpretability improvement of the synthesized image patterns~\cite{Yosinski:2015:ICML:AM,Nguyen:2016:ICML:AM-multifaceted,Nguyen:2016:NIPS:DGN-AM}.
	2) Besides the visualization of a single neuron, Mahendran \textit{et al.} revealed the CNN internal features in the layer level~\cite{Mahendran:2015:CVPR:Network-Inversion,Mahendran:2016:CVPR:Network-Inversion}. The \textit{Network Inversion} was proposed to reconstruct an input image based on multiple neurons' activation to illustrate a comprehensive feature map learned by each single CNN layer.
	3) Rather than reconstructing an input image for feature visualization, Zeiler \textit{et al.} proposed the \textit{Deconvolutional Neural Network based Visualization} (\textit{DeconvNet})~\cite{Zeiler:2014:ECCV:DeconvolNet}, which utilized the \textit{DeconvNet} framework to project the feature map to an image dimension directly.
	With the direct projecting, \textit{DeconvNet} can highlight what patterns in the input image activate the specific neurons and hence link the neurons and the meaning of input data directly.
	4) Recently, Zhou \textit{et al.}~\cite{Bau:2017:CVPR:Network-Dissection} proposed the \textit{Network Dissection based Visualization}, which interpreted the CNN in the semantic level.
	By referencing a heterogeneous image dataset -- \textit{Borden}, the \textit{Network Dissection} can effectively partition the input image into multiple sections with various semantic definitions.
	As the semantics directly represent the feature meanings, the neuron interpretability can be significantly enhanced.

This survey paper is expected to provide a comprehensive review of these representative CNN visualization methods, in terms of motivations, algorithms, and experiment results. We also discuss the practical applications of the CNN visualization, demonstrating the significance of the network interpretability in areas of network design, optimization, security enhancement, \textit{etc}.

The rest of the paper is organized as follows:
	In Section 2, we introduce the background knowledge of the CNN and visualization.
	In Sections 3$\sim$6, we describe the four aforementioned representative visualization methods, namely the \textit{Activation Maximization}, \textit{DeconvNet}, \textit{Network Inversion}, and \textit{Network Dissection} based visualization.
	In Section 7, we present several CNN visualization applications.
	In Section 8, the CNN visualization research potentials are discussed with the conclusion.

\section{Background}
In this section, we introduce the background knowledge of the CNN structure, algorithm, and CNN visualization.

\subsection{CNN structure}
\begin{figure}[b!]
	\begin{center}
		\includegraphics[width=4.8in, height=1.5in]{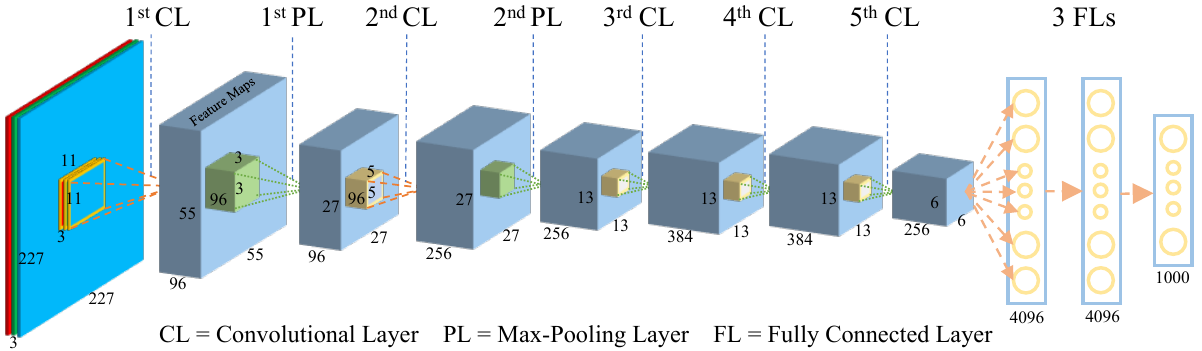}\\
		\caption{\textit{CaffeNet} architecture}\label{caffenet}
	\end{center}
\end{figure}
In machine learning, the CNN is a type of deep neural networks (DNNs), which has been widely used for computer vision related tasks.
	Fig.~\ref{caffenet} shows a representative CNN structure -- \textit{CaffeNet}~\cite{Jia:2014:caffe}, which is a replication of \textit{AlexNet} with 5 convolutional layers (CLs) and 2 max-pooling layers (PLs) followed by 3 fully-connected layers (FLs):

\textit{Convolutional Layer}:
	Fig.~\ref{caffenet} demonstrates the CNN structure, and the yellow blocks represent the convolutional filters -- neurons in the convolutional layers for feature extraction. These filters perform the convolution process to transform the input images or previous layer feature maps into the output feature maps, which are denoted as the blue blocks in Fig.~\ref{caffenet}.

\begin{figure}[t!]
	\begin{center}
		\includegraphics[width=4in, height=1.8in]{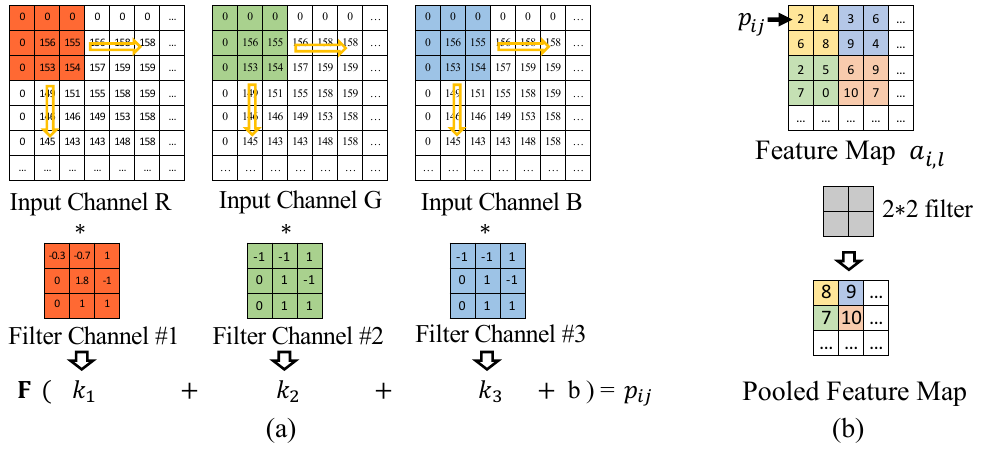}\\
		\caption{Convolutional and max-pooling process}\label{CP}
	\end{center}
\end{figure}

Fig.~\ref{CP} (a) shows the detailed convolutional process of the first CL.
	The convolutional filters in the first layer have three channels corresponding to the three RGB color dimensions of the raw input images.
	Each filter channel performs dot production in a small region of the input data to compose a color specific feature channel.
	This process is usually followed by an activation function $F$, usually \textit{relu}~\cite{Collobert:2011:relu}, which gives a summation of dot productions when positive or 0 otherwise.
	Hence, we can get a color comprehensive element $p_{ij}$ of the final rectified feature map.
	Based on the small region convolution, each filter is replicated across the entire input image. Multiple $p_{ij}$s would produce the final rectified feature map (or activation map) $a_{i,l}$.

The rectified feature maps represent the extracted features and would act as the inputs for the next CL.
	Mathematically, for the filter $i$ in layer $l$, This process can be viewed as:
\begin{equation}\label{convolution}
	a_{i,l+1} = F( \sum w_{i,l} a_{i,l} + b_{i,l}),
\end{equation}
where $w, b$ represents the weights and bias parameters respectively.

With such a process, the filters act as feature extractors from the original input image to learn the useful features for classification.

\textit{Pooling Layer}:
As shown in Fig.~\ref{caffenet}, after each CL, it's optimal to apply a PL on the output feature maps.
	As denoted by the green blocks, the pooling filters perform the down-sampling operation to reduce the data dimension of the input feature maps ($a_{i,l}$).
	Fig.~\ref{CP} (b) shows the max-pooling process, which is a widely adopted pooling method.
	The max-pooling is achieved by applying a $2 \times 2$ pooling window to select the maximal element of a $2 \times 2$ region of the input feature map with a stride of 2.
	This process aggressively reduces the spatial size of the feature maps and condense the extracted feature information.

Hence, the pooling layers contribute the CNN with fewer data redundancy and therefore less data processing workload.

\textit{Fully-connected Layer}:
In Fig.~\ref{caffenet}, each yellow circle represent one neuron in the FLs, which is connected to all the previous input feature maps.
	The FLs perform a comprehensive feature evaluation based on the features extracted by the CLs, and generate an N-dimensional probability vector, where N is the number of the classification targets.
	For example, in the digit classification task, N would be 10 for the digits of 0$\sim$9~\cite{mnist}.

After the final layer of FLs, a \textit{SoftMax} function is used to generate the final classification probability as defined in the Eq.~\ref{softmax}:
\begin{equation}\label{softmax}
	P_{i} = \frac{e^{a_{i}}}{ \sum_{N =1}^{10} e^{a_{n}}},
\end{equation}
where $a_{i}$ is $i^{th}$ neuron output in the final FL layer.
	This function normalizes the final FL layer output to a vector of values between zero and one, which gives a probability over all 10 classes.

By applying the above-mentioned hierarchical structured layers, CNNs transform the input image layer by layer from the original pixel values to the final probability vector $P$, in which the largest $P_{i}$ indicates the most predicted class.

\subsection{CNN algorithm}
The CNNs not only benefit from the deep hierarchical structure, but also the delicate learning algorithm~\cite{Lecun:1998:backpropgation}.
	The learning algorithm aims to minimize the training error between the predicted values and actual labels by updating the network parameters, which are quantified by the loss function.
	The training error can be viewed as:
\begin{equation}\label{loss}
	C(w, b) = \frac{1}{n} \sum_{n}^{i=1} L(w,b,x_{i},y_{i}),
\end{equation}
where $L(\cdot)$ represents the loss function, and $(x_{1},y_{1}),...(x_{n},y_{n})$ represent the training examples.
	During the learning process, a square loss is usually applied, then the loss function will be:
\begin{equation}\label{loss2}
	L(w,b,x_{i},y_{i}) = (y_{i} - f(w, b, x_{i}))^{2},
\end{equation}
	where $f(\cdot)$ indicates the predicted values calculated by the whole CNN:
\begin{equation}\label{f}
	f(w, b, x)= F( \sum w_{i, l} F (w_{i, l-1}F(...F(\sum w_{i,1} x_{i,0} + b_{i,0})...) + b_{i, l-1}) + b_{i, l}).
\end{equation}
In order to minimize the $C(w,b)$, a partial derivative  ${\partial C} / {\partial (w, b)}$ with respect to each weight $w$ and bias $b$ is calculated by backpropagating through all the layers in the CNN.
	The gradient descent method is utilized to iteratively update all parameter values.
	The update procedure for $w$ from iteration $j$ to $j + 1$ can be viewed as:
\begin{equation}\label{f}
	w_{j+1} = w_{j} - \eta \cdot \frac{\partial C(w; x, y)} {\partial w},
\end{equation}
where $\eta$ is the learning rate.
	Before the learning process, the parameters are usually randomly initialized~\cite{Glorot:2010:random_initial}.
	With the learning process, the convolutional filters become well configured to extract certain features.
	The features captured by convolutional filters can be demonstrated by visualization.

A lot of works have been proposed to optimize the structure and algorithm of the CNNs.
	For example, much deeper network structures have been investigated, such as \textit{VGG}, \textit{GoogleNet}, and \textit{ResNet}.
	At the same time, some regularization and optimization techniques have been applied, such as dropout~\cite{Srivastava:2014:dropout}, batch normalization~\cite{Ioffe:2015:batch}, momentum~\cite{Qian:1999:momentum}, and adagrad~\cite{Duchi:2011:adaptive}.
	As a result, CNNs have been well optimized and widely used in computer vision related tasks. However, the CNNs still suffer from high computational cost, slow training speed, and large training dataset requirement, which highly compromise the applicability and performance efficiency~\cite{Han:2015:deep-compression}.
	Hence, these weakness require more understanding about the CNN working mechanism to further optimize the CNN.

\subsection{CNN visualization mechanism}
\begin{figure}[htp]
	\begin{center}
		\includegraphics[width=4.7in, height=1.7in]{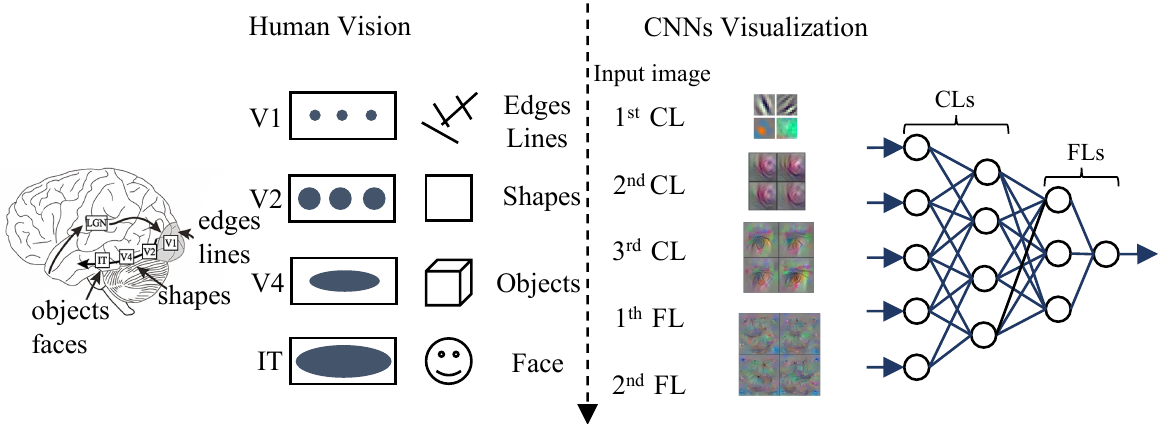}\\ \caption{Human vision and CNNs visualization}\label{visualization}
	\end{center}
\end{figure}
CNN visualization is a well utilized qualitative method to analysis the CNN working mechanism regarding the network interpretability.
	The interpretability is related to the ability of the human to understand the CNNs, which can be improved by demonstrating the internal features learned by CNNs.
	Visualization greatly helps to interpret the CNN's internal features, since it utilizes the human visual cortex system as a reference.

Fig.~\ref{visualization} shows how human visual cortex system process the visual information and how CNNs extract the features.
	As shown in the left part Fig.~\ref{visualization}, the human visual system processes the object features in a feed-forward and hierarchical approach through multiple visual neuron areas.
When humans recognize a face, the visual neurons with small receptive fields in the lower visual neuron area (\textit{e.g.} V1), are sensitive to basic visual features~\cite{crowding,Hubel:1959:Receptive,receptive}, such as edges and lines.
	In the higher visual neuron areas (\textit{e.g.} V2 and V4), the visual neurons have larger receptive fields, and are sensitive to complex features, such as shapes and objects.
	In the visual neuron area of IT, the visual neurons have the largest and most comprehensive receptive fields, therefore they are sensitive to the entire face.

For CNN interpretability study, researchers found the similar feature representation through the CNNs visualization as shown in the right part of Fig.~\ref{visualization}.
	Typically, the CNN feature extraction starts with small features such as edges and colored blobs in the first convolutional layer.
	Then the feature extraction progresses into general shapes and partial objects with deeper layers, and ends in a final classification with the fully-connected layers.
	By comparing the functionalities of brain visual neurons' receptive fields to the CNN's neurons~\cite{Kruger:2013:visual-hierarchies}, visualization illustrates the functionalities of each component in the CNNs.

\subsection{CNNs visualization methods}
\begin{figure}[b]
	\begin{center}
		\includegraphics[width=5in, height=1.6in]{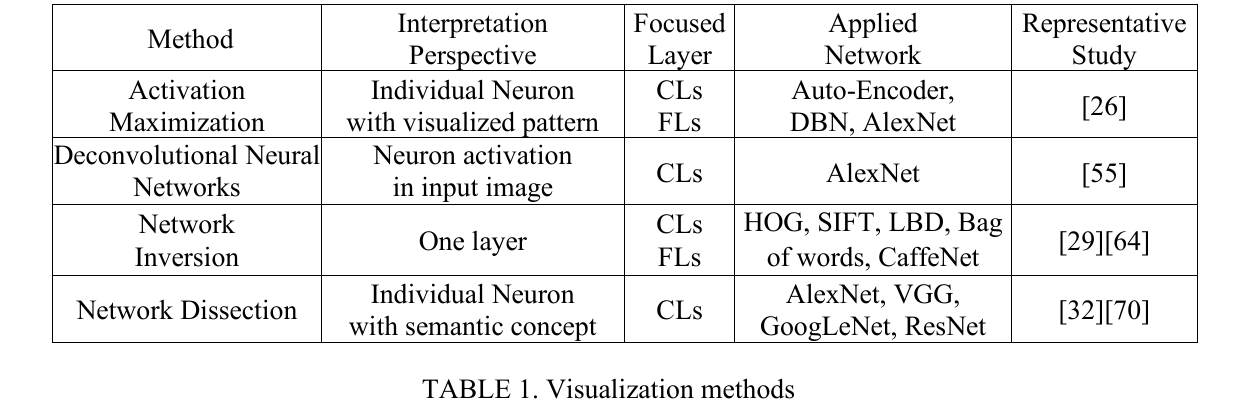}
	\end{center}
\end{figure}
The similarity between how the human vision system and CNN recognizes image inspired research to work on interpreting CNNs, and lots of CNN visualization works of the learned features have been widely discussed.
	In the early research stage, the visualization mainly focused on the low-level features~\cite{Poultney:2007:efficient,Lee:2008:sparse,Lee:2009:convolutional}.
	With the rapid developments and implementations of CNNs, the visualization has been extended to interpret the overall working mechanism of CNNs.
	In Table 1, we give a brief review of four representative visualization methods, namely \textit{Activation Maximization}, \textit{Deconvolutional Networks} (\textit{DeconNet}), \textit{Network Inversion}, and \textit{Network Dissection}:\vspace*{3pt}

$\bullet$ In \textit{Activation Maximization}, a visualized input image pattern is synthesized to illustrate a specific neuron's max stimulus in each layer;\vspace*{3pt}

$\bullet$ \textit{DeconNet} utilizes an inversed CNN structure, which is composed deconvolutional and unpooling layers, to find the image pattern in the original input image for a specific neuron activation;\vspace*{3pt}

$\bullet$ \textit{Network Inversion} reconstructs an input image based on the original image from a specific layer's feature maps, which reveals what image information is preserved in that layer;\vspace*{3pt}

$\bullet$ \textit{Network Dissection} describes neurons as visual semantic detectors, which can match six kinds of semantic concepts (\textit{e.g.} scene, object, part, material, texture, and color).\vspace*{3pt}

To compare these methods directly, we summarize the overview, algorithms, and visualization results of these methods:
	1) The overview summarizes the history and represent works in this line of work.
	2) The algorithms explain how this method works for the CNNs visualization.
	3) The visualization results provide a comprehensive understanding how CNNs extract features.
\section{Visualization by activation maximization}
\textit{\vspace{1mm}\\\-\hspace{0.5cm}Synthesize an input pattern image that can maximize a specific neuron's activation in arbitrary layers.}

\subsection{The overview}
\textit{Activation Maximization} (AM) is proposed to visualize the preferred inputs of neurons in each layer.
	The preferred input can indicate what features of a neuron has learned.
	The learned feature is represented by a synthesized input pattern that can cause maximal activation of a neuron.
	In order to synthesize such an input pattern, each pixel of the CNN's input is iteratively changed to maximize the activation of the neuron.

The idea behind the AM is intuitive, and the fundamental algorithm was proposed by Erhan \textit{et al}. in 2009~\cite{Erhan:2009:ICML:AM}.
	They visualized the preferred input patterns for the hidden neurons in the Deep Belief Net~\cite{Hinton:2006:DBNs} and the Stacked Denoising Auto-Encoder~\cite{Vincent:2010:SDAE} learned from the MNIST digit dataset~\cite{mnist}.
	Later, Simonyan \textit{et al}. utilized this method to maximize the activation of neurons in the last layer of CNNs~\cite{Simonyan:2013:ICLR:AM}.
	Google also has synthesized similar visualized patterns for their inception network~\cite{Mordvintsev:2015:Google-dream}.
	Yosinksi \textit{et al.} further applied the AM in a large scale, which visualized the arbitrary neurons in all layers of a CNN~\cite{Yosinski:2015:ICML:AM}.
	Recently, a lot of optimization works have followed this idea to improve the interpretability and diversity of the visualized patterns~\cite{Nguyen:2016:ICML:AM-multifaceted,Nguyen:2016:NIPS:DGN-AM}.
	With all these works, the AM has demonstrated great capability to interpret the interests of neurons and identify the hierarchical features learned by CNNs.

\subsection{The algorithm}
In this section, the fundamental algorithm of the AM is presented.
	Then, another optimized AM algorithm is discussed, which dramatically improves the interpretability of visualized patterns by utilizing a deep generator network~\cite{Nguyen:2016:NIPS:DGN-AM}.

\subsubsection{Activation maximization}
The fundamental algorithm of the AM can be\break viewed as synthesizing a pattern image $x^{*}$, which maximizes the activation of a target neuron:
\begin{equation}\label{Activation Maximization}
	x^{*} = \mathop{\argmax}_{x} {a_{i,l}(\theta , x)},
\end{equation}
where $\theta$ denotes the network parameter sets (weight and bias).

This process can be divided into four steps:

(1) An image $x = x_{0}$ with random pixel values is set to be the input to the activation computation.

(2) The gradients with respect to the noise image $\frac{\partial a_{i,l}}{ \partial x}$ are computed by using backpropagation, while the parameters of this CNN are fixed.

(3) Each pixel of the noise image is changed iteratively to maximize the activation of the neuron, which is guided by the direction of the gradient $\frac{\partial a_{i,l}}{ \partial x}$.
Every single iteration in this process applies the update:
\begin{equation}\label{gradient ascent}
	x \leftarrow x + \eta \cdot \frac{\partial a_{i,l}(\theta , x)}{\partial x},
\end{equation}
where $\eta$ denotes the gradient ascent step size.

(4) This process terminates at a specific pattern image $x^{*}$,  when the image without any noise.
This pattern is seen as preferred input for this neuron~\cite{Yosinski:2015:ICML:AM}.

Typically, we are supposed to use the unnormalized activation $a_{i} (\theta , x)$ of class $c$ in the final CNN layer of this visualization network , rather than the probability returned by the \textit{SoftMax} in Eq. ~\ref{softmax}.
	Because the \textit{SoftMax} normalize the final layer output to a vector of values between zero and one, the maximization of the class probability can be achieved by minimizing the probability of other classes.
	This method can be applied to any kinds of CNNs as long as we can compute the aforementioned gradients of the image pattern.

\subsubsection{Activation maximization with regulation}
However, the AM method has a considerable shortcoming: as the CNN becoming deeper, the visualized patterns in higher layers are usually tend to be unrealistic and uninterpretable.
In order to find the human-interpretable patterns, many regularization methods have been experimentally shown to improve the interpretability of the patterns.

A regularization parameter of $\lambda (x)$ is usually introduced to bias the visualized pattern image:
\begin{equation}\label{Activation Maximization}
	x^{*} = \mathop{\argmax}_{x}( {a_{i,l}(\theta , x) - \lambda (x)}).
\end{equation}
Different methods are adopted to implemented the $\lambda (x)$, such as $\ell_2$ decay, Gaussian blur, mean image initialization, and clipping pixels with very small absolute value~\cite{Simonyan:2013:ICLR:AM,Yosinski:2015:ICML:AM,Wei:2015:intra-class,Nguyen:2016:ICML:AM-multifaceted}.
For example, the $\ell_2$ decay tends to prevent a small number of extreme pixel values from dominating the visualized patterns.
The Gaussian blur penalize high frequency information in the visualized patterns, and the contribution of a pixel is measured as how much the activation increases or decreases when the pixel is set to zero.
Each of these regularization methods can be applied to the AM individually or cooperatively.
In the Section \ref{hidden_layer}, we shows the bias patterns by applying these regularization methods to improve the interpretability.

\subsubsection{Activation maximization with generator networks}
Instead of utilizing regularizer $\lambda (x)$ to bias the visualized, Nguyen \textit{et al.}~\cite{Nguyen:2016:NIPS:DGN-AM} utilized a image generator network~\cite{Dosovitskiy:2016:generating,Goodfellow:2014:NIPS:GAN} to to replace the iterative random pixel tuning, which maximize the activation of the neurons in the final CNN layer.
	The synthesized pattern image by the generator is more close to the realistic image, which greatly improves the interpretability of the visualized patterns.

Recently, most of the generator networks related works are based on Generative Adversarial Networks (GAN)~\cite{Goodfellow:2014:NIPS:GAN}.
	GANs can learn to mimic any distribution of data and generate realistic data samples, such as image, music, and speech, which is featured with a complementary composition of two neural networks:
	One generative network takes noise as input and aim to generate realistic data samples.
	Another discriminator network receives the generated data samples from the output of the generative network and the real data samples from the training data sets, which aim to distinguishes between the two sources.
	The goal of generative network is to generate passable data samples, to lie without being distinguished by the discriminator network.
	The goal of the discriminator is to identify images coming from the generative network as fake.
	After fine training of both the networks, the GAN eventually achieves a balance, where the discriminator can hardly distinguish generated data samples from real data samples.
	In such a case, we can claim that the generative network has achieved an optimal capability in generating realistic samples.
So far GANs have particularly produced excellent results in image data, and primarily been used to generate samples of realistic images~\cite{Denton:2015:GAN1,Ledig:2016:GAN2,Arjovsky:2017:wasserstein}.

Benefit from the success of GANs, the generative network is utilized to overcome the aforementioned shortcoming of AM that the visualized patterns in higher layers are usually tend to be unrealistic and uninterpretable.
	The generative network is utilized to generate or synthesize the pattern image that maximize the activation of the selected neuron $a_{i,l}$ in the final layer~
	This method is called ~\textit{Deep Generative Network Activation Maximization} (DGN-AM).

The DGN-AM implemention can be view as:
\begin{equation}\label{DGN}
	x^{*} = \mathop{\argmax}_{x}{(a_{i,l}(\theta, G(x))- \lambda (x))},
\end{equation}
where $G$ indicates the generative network that takes the noise image as input.
	It can synthesize the pattern image that causes high activation of the target neuron $a_{i,l}$.
	In ~\cite{Nguyen:2016:NIPS:DGN-AM}, the author found that the $\ell_2$ regularization with small degree helps to generate more human-interpretable patterns.
	In the Section \ref{AM_Output_3}, we compare the pattern image synthesized by the AM and DGN-AM.

\subsection{Experiments with activation maximization}
In this section, the experiments of AM on \textit{CaffeNet} trained with \textit{ImageNet} dataset are demonstrated, which shows what features have been learned by each neuron.
	The first layer neurons can be visualized by directly mapping each filter matrix to an RGB pattern image, hence, the first layer visualization by AM and the direct mapping method are evaluated to show the effectiveness of AM.
Then the hidden layer visualization shows the abundant and hierarchical features learned by different layer neurons.
Finally the final layer visualization by AM and DGN-AM are compared to emphasize the improvement of interpretability by the DGN-AM.

\subsubsection{First layer visualization}
As aforementioned, the AM has more distinguishable performance on the early network layers, hence, we first evaluate the first layer visualization with AM.
\begin{figure}[htp]
\begin{center}
	\includegraphics[width=3in, height=2.2in]{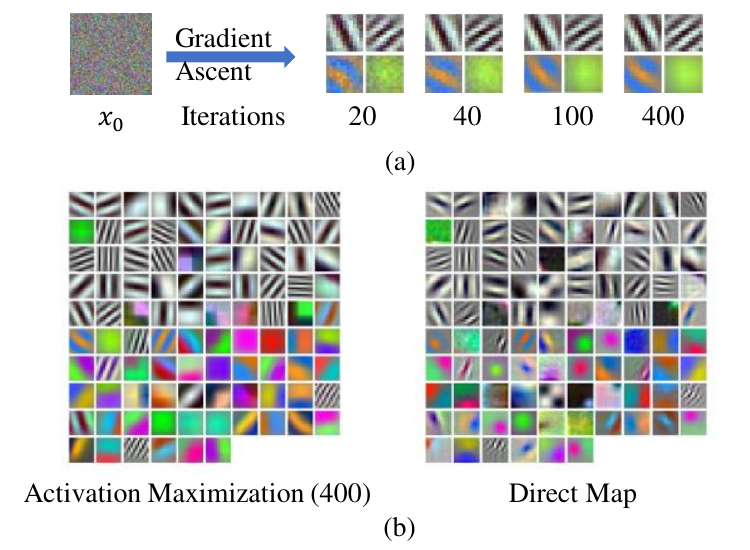}\\
	\caption{First layer of \textit{CaffeNet} visualized by \textit{Activation Maximization}}\label{am_first}
	\end{center}
\end{figure}

Fig.~\ref{am_first} (a) shows several visualization results of synthesizing the patterns for four different neurons.
	This process uses the gradient $\frac{\partial a_{i,l}}{\partial x}$ to tweak the input noise iteratively, which increases the activation of neuron $i$.
	After 400 iterations, we can get relatively smooth visualized patterns, which indicates a nicely converged network is trained.

Fig.~\ref{am_first} (b) shows the visualized patterns synthesized by the AM and direct mapping method.
	As we can see, most of the visualized patterns synthesized by the AM are almost the same as the corresponding direct mapped patterns.
	The visualized patterns are clustered into two groups:
	1) the colorful patterns indicate the corresponding neurons greatly sensitive to color components in the under-test images;
	2) the black-and-white patterns indicate the corresponding neurons greatly sensitive to shape information.
	In addition, through comparison with the direct map method, the AM can reveal the preferred inputs of each neuron accurately.

This interesting finding reveals that the CNNs attempt to imitate the human visual cortex system, which the neurons in the lower visual area are sensitive to basic patterns, such as colors, edges, and lines.
\begin{figure}[htp]
	\begin{center}
		\includegraphics[width=5in, height=3.7in]{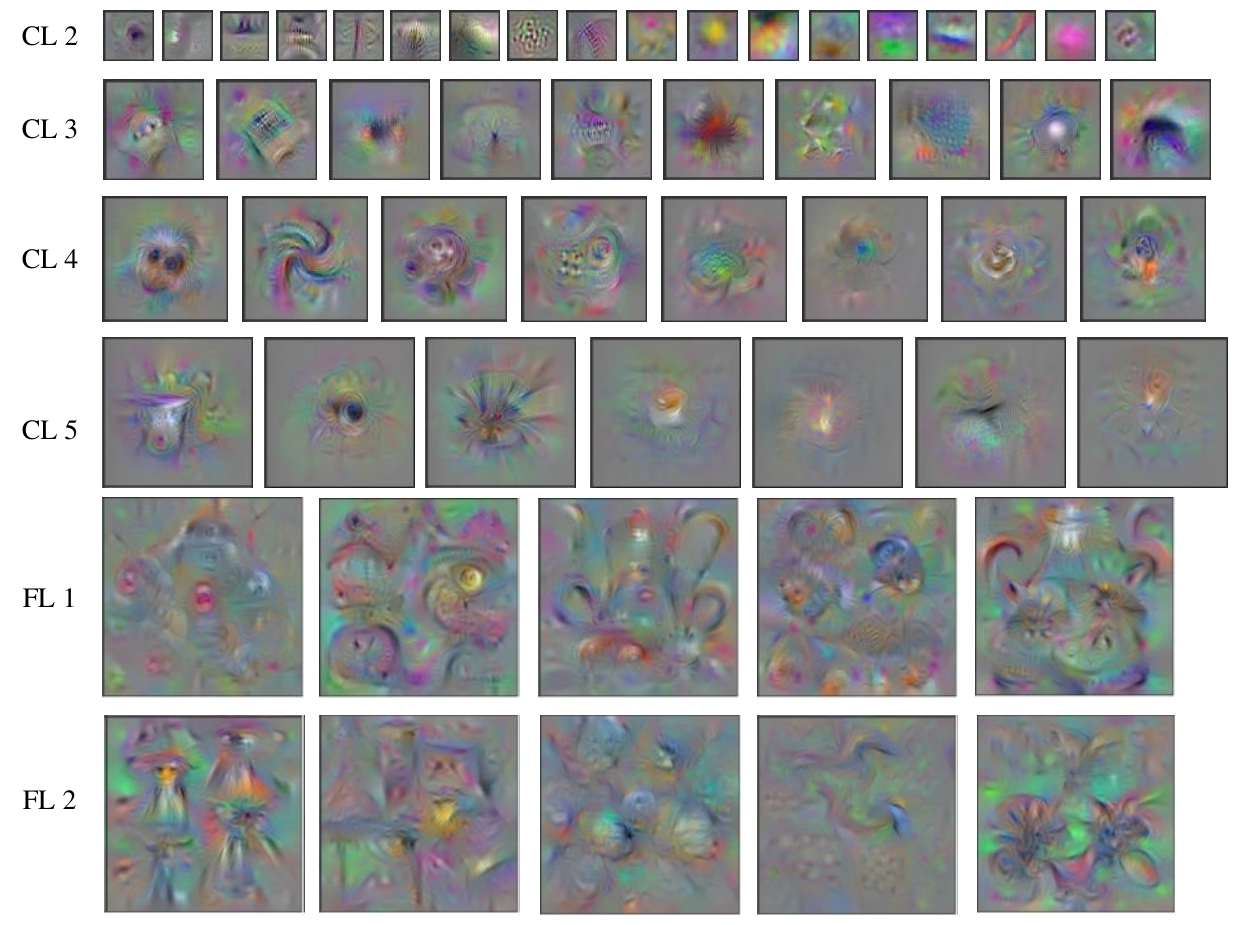}\\
		\caption{Hidden layers of \textit{CaffeNet} visualization by \textit{Activation Maximization}.}\label{am_hidden}  Adapted from ``Understanding Neural Networks Through Deep Visualization,'' by J. Yosinski, 2015.
	\end{center}
\end{figure}

\subsubsection{Hidden layers visualization} \label{hidden_layer}
Beyond the first layer, the neurons in the following layers gradually learn to extract feature hierarchically.

Fig.~\ref{am_hidden} shows the visualization of 6 hidden layers from the second convolutional layer (CL 2) to the second fully connected layer (FL 2) in each row.
	Several neurons in each layer are randomly selected as our AM test targets.
	We observed that: 1) Some important patterns are visualized, such as edges (CL 2-4), faces (CL 4-1), wheels (CL 4-2), bottles (CL 5-1), eyes (CL 5-2), \textit{etc}., which demonstrate the abundant features learned by the neurons.
	2) Meanwhile, not all the visualized patterns are interpretable even with multiple regularization methods are applied.
	3) The complexity and variation of the visualized patterns are increasing from lower layers to higher layers, which indicates that increasingly invariant features are learned by the neurons.
	4) From the CL 5 to FLs, we can find there is a large pattern variation increment, which could indicate the FLs provide a more comprehensive feature evaluation.
\begin{figure}[b]
	\begin{center}
		\includegraphics[width=5in, height=2in]{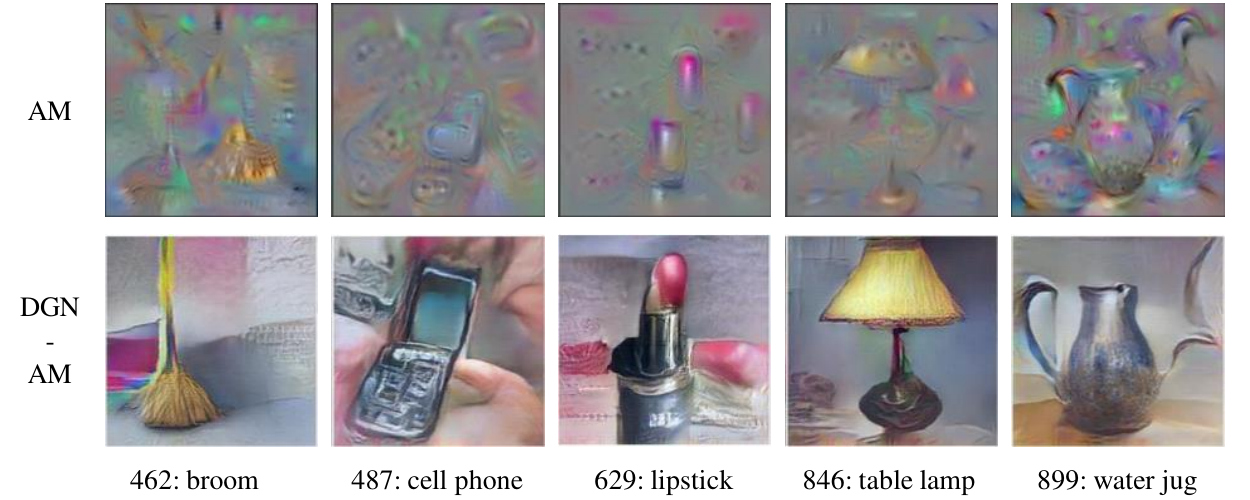}\\
		\caption{Output layer of \textit{CaffeNet} visualized by \textit{Activation Maximization}.}\label{am_output}
	\end{center}
\end{figure}

\subsubsection{Final layer visualization} \label{AM_Output_3}
The final layer of the \textit{CaffeNet} contains 1000 neurons, which corresponding to the 1000 classes in the \textit{ImageNet} dataset.
	Five different neurons are selected to show the visualized pattern image.

Fig.~\ref{am_output} compares the visualized patterns synthesized by the AM and the DGN-AM for five different classes in FL 3.
	For the AM shown in the first row of Fig.~\ref{am_output}, although we can guess which class the visualized patterns represents, there are multiple duplicate and vague objects in the each visualized pattern, such as three lipsticks in the third column (AM-3) and the images are far from being photo-realistic.
	For the DGN-AM shown in the second row of Fig.~\ref{am_output}, by utilizing the generator networks, the DGN-AM greatly improves the images quality in terms of color and texture.
	Because the fully connected layer contain information from all areas of the image and the generator network provides a strong biases toward realistic visualizations.

Through the final layer visualization, we can clearly see what objects combinations could affect the CNN classification decision.
	For example, if the CNN classifies an image of a cell phone held in a human hand as a cell phone, it is unclear if the classification decision is affected by the human hand.
	Through the visualization, we can see there is a cell phone and a human hand in the cell phone class, which are shown in the second row and third column.
	In this case, the visualization shows the CNN has learned to detect both object information in one image.

\subsection{The summary}
As the most intuitive visualization method, the AM reveals that CNNs learn to detect the important features such as faces, wheels, and bottles without our specification.
	At the same time, CNNs attempt to mimic the hierarchical organization of the visual cortex, and then successfully build up the hierarchical feature extraction.
	In addition, this visualization method suggests that the individual neurons extract features in a more local manner rather than distributed, which each neuron correspond to a specific pattern.
\section{Visualization by deconvolutional network}
\textit{\vspace{1mm}\\\-\hspace{0.5cm}Find the selective pattern from a given input image that activate a specific neuron in the convolutional layers.}

\subsection{The overview}
While the \textit{Activation Maximization} interprets the CNNs from the perspective of the neurons, the \textit{Deconvolutional Network} (\textit{DeconvNet}) based CNN visualization explains the CNNs from the perspective of the input image.
	It finds the selective patterns from the input image that activate a specific neuron in the convolutional layers.
	The patterns are reconstructed by projecting the low-dimension neurons' feature maps back to the image dimension.
	This projection process is implemented by a \textit{DeconvNet} structure, which contains deconvolutional layers and unpooling layers, performing the inversed computation of the convolutional and pooling layers.
	Rather than purely analyzing the neurons' interests, the \textit{DeconvNet} based visualization demonstrates a straightforward feature analysis in an image level.

The research related to the \textit{DeconvNet} structure is mainly led by Zeiler~\textit{et al}.
	In~\cite{Zeiler:2010:CVPR:DeconvolNet}, they first proposed the \textit{DeconvNet} structure aiming to capture certain general features for reconstructing the natural image
	by projecting a highly diverse set of low-dimension feature maps to high dimension.
	Later in~\cite{Zeiler:2011:ICCV:DeconvolNet}, they utilized the \textit{DeconvNet} structure to decompose an image hierarchically, which could capture the image information at all scales, from low-level edges to high-level object parts.
	Eventually, they applied the \textit{DeconvNet} structure for CNN visualization by interpreting CNN hidden features~\cite{Zeiler:2014:ECCV:DeconvolNet}, which made it become an effective method to visualize the CNNs.

\subsection{The algorithm}
The \textit{DeconvNet} is an effective method to visualize the CNNs, we will explain the \textit{DeconvNet} based visualization in terms of \textit{DeconvNet} structure and the visualization process in this section.

\subsubsection{\textit{DeconvNet} structure}
The~\textit{DeconvNet} provides a continuous path, which\break projects the low-dimension pooled feature map back to the image dimension.
	Typically, there are reversed convolutional layers (namely the deconvolutional layer), reversed rectification layers, and reversed max-pooling layers (namely unpooling layer) in the ~\textit{DeconvNet} structure.
	A typical \textit{DeconvNet} structure is shown in Fig.~\ref{deconvnet}.
	In Fig.~\ref{deconvnet}(a), the \textit{DeconvNet} serves as a reversed process of CNN, which is composed of the reversed layers corresponding to the layers in CNNs.

Each layer of the~\textit{DeconvNet} is defined as follows:
\begin{figure}[t]
	\begin{center}
		\includegraphics[width=5in, height=2.3in]{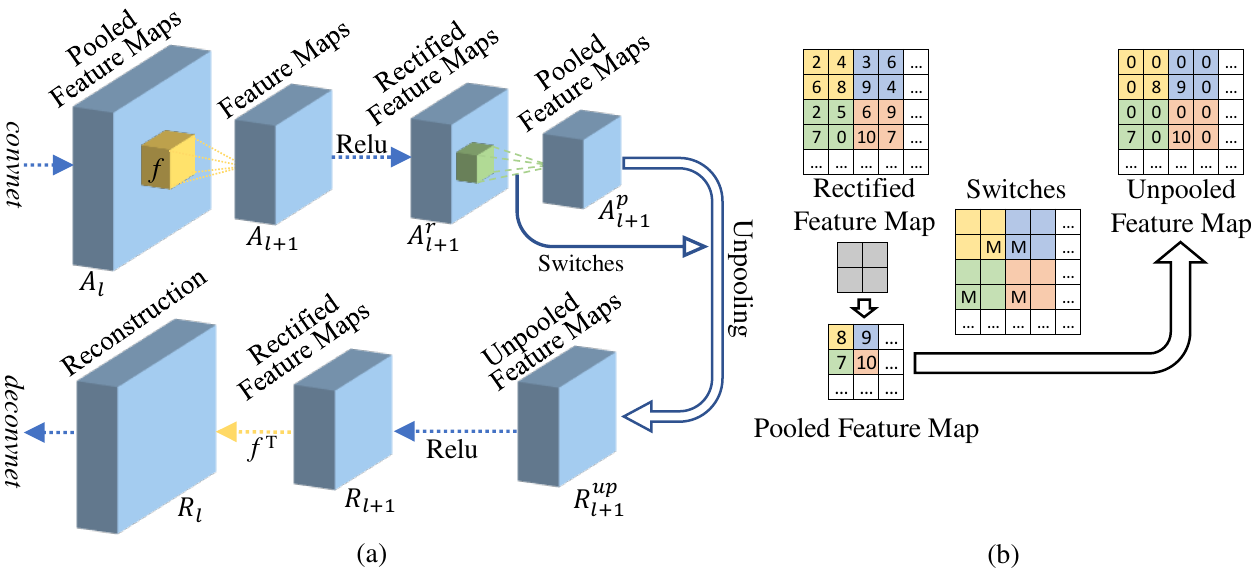}\\
		\caption{The structure of the \textit{Deconvolutional Network}}\label{deconvnet}
	\end{center}
\end{figure}

\textit{Reversed Convolution/Deconvolutional Layer}:
To explain the deconvolutional layer, we first take a look at the convolutional layers as shown in the top Fig.~\ref{deconvnet} (a).
	The convolutional layer transforms the input feature maps into the output feature maps described in Eq.~\ref{convolution}: $a_{i,l+1} = F( \sum w_{i,l} a_{i,l} + b_{i,l})$, where the $w, b$ is the filter parameters and $F$ is the \textit{relu} activation function.
	We combine the convolutional and summing operations of layer $l$ into a single filter matrix $f_{l}$ and convert the multiple feature maps $a_{i,l}$ into a single feature vector $A_{l}$:
\begin{equation}\label{convolutional layer}
A_{l+1} = A_{l}*f_{l}.
\end{equation}
	After appling the \textit{relu} function, the rectified feature maps $A^{r}_{l}$ is produced.

While in the deconvolutional operation, the reversed convolutional layer, namely the deconvolutional layer, uses the transposed versions of the same convolutional filters to perform the convolutional operations.
	The deconvolution process can be viewed as:
\begin{equation}\label{deconvolutional layer}
R_{l} = R_{l+1} * f_{l}^{T}.
\end{equation}
	The $f_{l}^{T}$ is the transposed versions of the convolutional filters, which is flipped from the filters $f$ horizontally and vertically.
	The $R_{l}$ indicates the rectified feature maps in the \textit{DeconvNet}, which is convolved with the $f_{l}^{T}$.

\textit{Reversed Rectification Layer}:
The CNNs usually use the \textit{relu} activation function, which rectifies the feature maps thus ensuring the feature maps are always positive.
	The feature maps of deconvolutional layer are also ensured to be positive in reconstruction by passing the unpooled feature maps $R^{up}$ through a \textit{relu} function.

\textit{Reversed Max-pooling/Unpooling Layer}:
The reversed max-pooling process in a \textit{DeconvNet} is implemented by the unpooling layer.
	Fig.~\ref{deconvnet} (b) shows the unpooling process in detail:
	In order to implement the reversed operation of max-pooling, which performs the downsampling operation on the rectified feature maps $A^{r}_{l+1}$, the unpooling layer transform the pooled feature maps to the unpooled feature maps.

During the max-pooling operation, the positions of maximal values within each pooling window are recorded in switch variables.
The switches first specify the position of which elements in the rectified feature map are copied into the pooled feature map, then mark them as \textit{M} in the switches.
	These switches variables are used in the unpooling operation to place each maximal value back to its original pooled location.
Due to the dimension gap, certain amount of locations are inevitable constructed without certain information, therefore these locations are usually filled by zero for compensation.

\subsubsection{Visualization process}
Based on the reversed structure formed by those layers, the \textit{DeconvNet} can be well utilized to visualize the CNNs.
The visualization process can be described as follows:

(1) All neurons' feature maps can be captured when a specific input image is processed through the CNN.

(2) The feature map of the target neuron for visualization is selected while all other neurons' feature maps are set to zeros.

(3) In order to obtain the visualized pattern, the target neuron's feature map is projected back to the image dimension through the \textit{DeconvNet}.

(4) To visualize all the neurons, this process is applied to all neurons repeatedly and obtain a set of corresponding pattern images for CNN visualization.

These visualized patterns indicate which pixels or features in the input image contribute to the activation of the neuron, and it also can be used to examine the CNN design shortcomings.
	In the next section, these visualized patterns will be demonstrated with practical experiements.

\subsection{Experiments with \textit{DeconvNet} based visualization}
In this section, the experiments of \textit{DeconvNet} based visualization on \textit{CaffeNet} trained with \textit{ImageNet} dataset are demonstrated, to show what features have been learned by each neuron.
	In addition, this method can be used as an efficient tools for network analysis, optimization and training monitoring.

\subsubsection{Convolutional layer visualization}
\begin{figure}[b]
	\begin{center}
		\includegraphics[width=3.8in, height=1.5in]{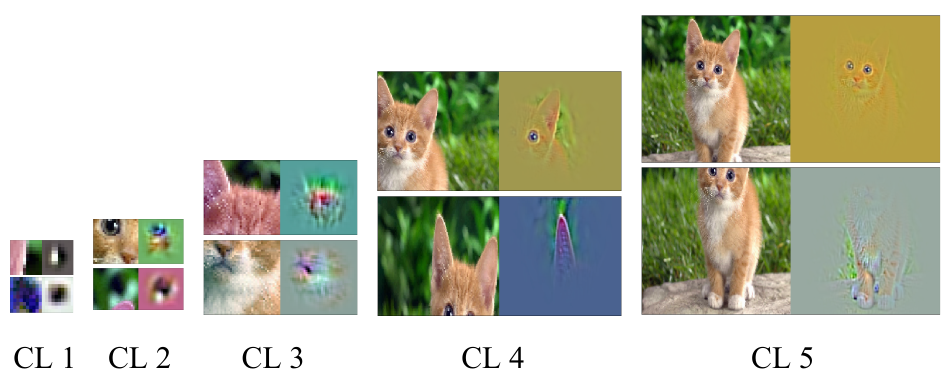}\\
		\caption{\textit{CaffeNet} visualized by \textit{DeconvNet}}\label{deconv_visualization}
	\end{center}
\end{figure}
As aforementioned, the \textit{DeconvNet} structure provides a continuous path back to the image dimension, which can highlight specific pattern in the input image that excites a neuron.

Fig.~\ref{deconv_visualization} shows the \textit{DeconvNet} based visualization examples with 5 convolutional layers of the \textit{CaffeNet} from CL1 to CL5.
	In each layer, we randomly select two neurons' visualized patterns comparing to the corresponding local sections in the original image.
	From these examples, we can tell that:
	Each individual neuron extracts feature in a more local manner, where different neurons in one layer are responsible for different patterns, such as mouth, eyes, and ears.

Lower layers (CL1, CL 2) capture the small edges, corners, and parts.
	CL3 has more complex invariance, capturing similar textures such as mesh patterns.
	Higher layers (CL4, CL5) are more class-specific, which show the almost entire objects.
	Compared to the~\textit{Activation Maximization}, the \textit{DeconvNet} based visualization can provide much more explicit and straightforward patterns.

\subsubsection{\textit{DeconvNet} based visualization for network analysis and optimization}
\begin{figure}[b]
	\begin{center}
		\includegraphics[width=4.6 in, height=2in]{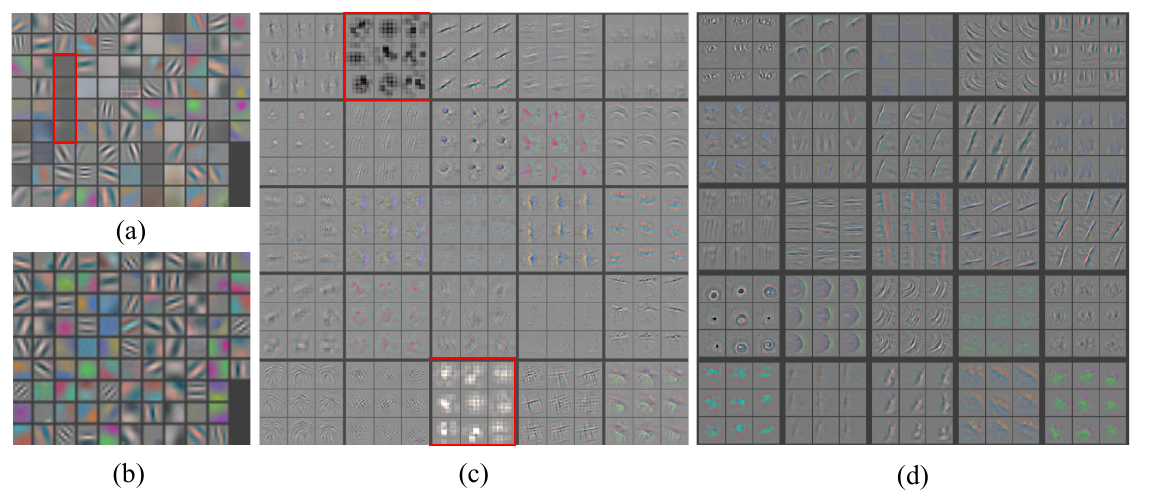}\\
		\caption{First and second layer visualization of ~\textit{AlexNet} and ~\textit{ZFNet}.}\label{deconv_visualization3} Adapted from ``Visualizing and Understanding Convolutional Networks,'' by M.D. Zeiler, 2014.
	\end{center}
\end{figure}
Beside the convolutional layer visualization for interpretation analysis, \textit{DeconvNet} can be also used to examine the CNN design for further optimization.

Fig.~\ref{deconv_visualization3} (a) and (c) show the visualization of the first and second layers from ~\textit{AlexNet}.
	We can find that:
	1) There are some ``dead" neurons without any specific patterns (indicated in pure gray color) in the first layer, which means they have no activation for the inputs.
	This could be a symptom of high learning rates or not good weights initialization.
	2) The second layer visualization shows aliasing artifacts, highlighting by the red rectangles.
	This could be caused by the large stride used in the first-layer convolutions.

These findings from the visualization can be well applied to the CNN optimization.
	Hence, Zeiler~\textit{et al.} proposed \textit{ZFNet}, which reduced the first layer filter size and shrink the convolutional stride of~\textit{AlexNet} to retain much more features in the first two convolutional layers.

The improvement introduced by \textit{ZFNet} is demonstrated in Fig.~\ref{deconv_visualization3} (b) and (d), which shows the visualizations of the first and second layers of \textit{ZFNet}.
	We can see that the patterns in the first layer become more distinctive, and the patterns in the second layer have no aliasing artifacts. Hence, the visualization can be effectively applied in CNN analysis and further optimization.

\subsubsection{\textit{DeconvNet} based visualization for training monitoring}
\begin{figure}[htp]
	\begin{center}
		\includegraphics[width=4.9 in, height=1.1in]{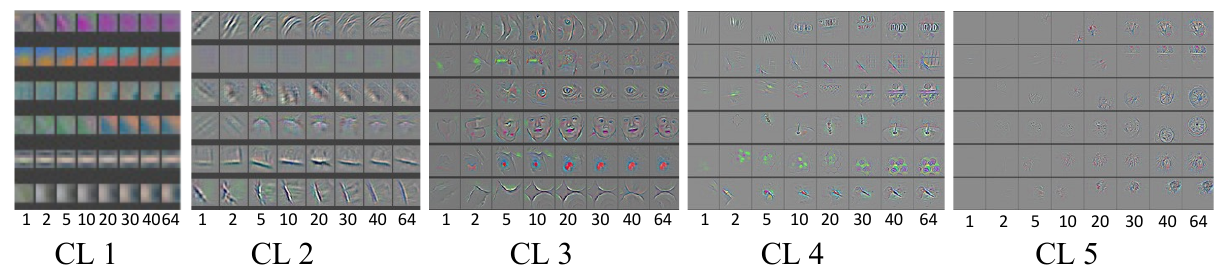}\\
		\caption{Feature evolution during training~\textit{ZFNet}}\label{deconv_visualization2}. Adapted from ``Visualizing and Understanding Convolutional Networks,'' by M.D. Zeiler, 2014.
	\end{center}
\end{figure}

Besides the CNN network optimization, the interpretability analysis can also help to monitor the CNN training process for better training efficiency.

Fig.~\ref{deconv_visualization2} shows the visualized pattens during the training the ~\textit{ZFNet}.
Each row indicates different neurons in the convolutional layers.
A randomly chosen subset of visualized patterns at different training epoch are shown in each column.
	We can find that: 1) In each row, the color contrast is artificially enhanced as the training process.
	2) The lower layers (CL1, CL2) converge quickly, since distinguished patterns appear within a few epochs.
	3) However, the distinguished patterns appear after a considerable number of epochs in the upper layers (CL4, CL5), which means these layers need to be trained until fully converged.

	Additionally, if noisy patterns are observed in the training process, that could indicate that the network hasn't been trained long enough, or low regularization strength that may result in overfitting.
	By visualizing features at several time points during training, we can find the design shortcomings and adjust the network parameters in time.
	In general, the visualization of training process is an effective way to monitor and evaluate the training statues.

\subsection{The summary}
The \textit{DeconvNet} highlights which selected patterns in the input image contribute to the activation of a neuron in a more interpretable manner.
	Additionally, this method can be used to examine the problems with the CNNs for optimization.
	And the training monitoring could provide the CNN research with a better criteria when adjusting the training configuration and stopping training.

However, both methods  of AM and \textit{DeconvNet} visualize the CNN in the neuron level, lacking a comprehensive perspective from higher structure, such as layer and whole network.
	In the following sections, we will further discuss high-level CNN visualization methods, which interpret each individual layer and visualize the information captured by the set of neurons in a layer as a whole.

\section{Visualization by network inversion}
\textit{\vspace{1mm}\\\-\hspace{0.5cm}Reconstruct an image from all the neurons' feature maps in an arbitrary layer to highlight the comprehensive CNN layer-level feature for  a given input image.}

\subsection{The overview}
Different from the activation from a single network neuron, the layer-level activation will reveal a comprehensive feature representation, which is composed of the all neuron activation patterns inside a layer.
	Hence, different form the aforementioned visualization methods, which visualize the CNN from a single neuron's activation, the \textit{Network Inversion} based visualization can be used to analysis the activation information from a layer level perspective.

Before the \textit{Network Inversion} is applied to visualize the CNNs, the fundamental idea of \textit{Network Inversion} was proposed to study the traditional computer vision representation, such as the Histogram of Oriented Gradients (HOG)~\cite{Dalal:2005:CVPR:HOG,Felzenszwalb:2010:Object_HOG,voc-release5}, the Scale Invariant Feature Transform (SIFT)~\cite{Lowe:2004:IJCV:SIFT}, the Local Binary Descriptors (LBD)~\cite{D:2012:LBD}, and the Bag of Visual Words Descriptors~\cite{Csurka:2004:ECCV:bags,Sivic:2003:ICCV:bags}.
	Later, two variants of the \textit{Network Inversion} were proposed for CNN visualization~\cite{Mahendran:2015:CVPR:Network-Inversion,Mahendran:2016:CVPR:Network-Inversion,Dosovitskiy:2016:CVPR:Network-Inversion}:

(1) Regularizer based \textit{Network Inversion}: It is proposed by Mahendran~\textit{et al}., which reconstructs the image from each layer by using gradient descent approach and a regularization term~\cite{Mahendran:2015:CVPR:Network-Inversion,Mahendran:2016:CVPR:Network-Inversion}.

(2) \textit{UpconvNet} based \textit{Network Inversion}: It is proposed by Dosovitskiy~\textit{et al}.~\cite{Dosovitskiy:2016:CVPR:Network-Inversion,Dosovitskiy:2015:up-convolutional}, which reconstructs the image by training a dedicated \textit{Up-convolutional Neural Network} (\textit{UpconvNet}).

Overall, the main goal of both algorithms is to reconstruct the original input image from one whole layer's feature maps' specific activation.
	The Regularizer based \textit{Network Inversion} is easier to be implemented, since it does not require to train an extra dedicated network.
	While the \textit{UpconvNet} based \textit{Network Inversion} can visualize more existent information in higher layers with an extra dedicated network and significantly more computational cost.

\subsection{The algorithm}
In this section, we compared the aforementioned two \textit{Network Inversion based Visualization} methods regarding the network structure and the learning algorithm.

\begin{figure}[htp]
\begin{center}
  \includegraphics[width=5in, height=2.2in]{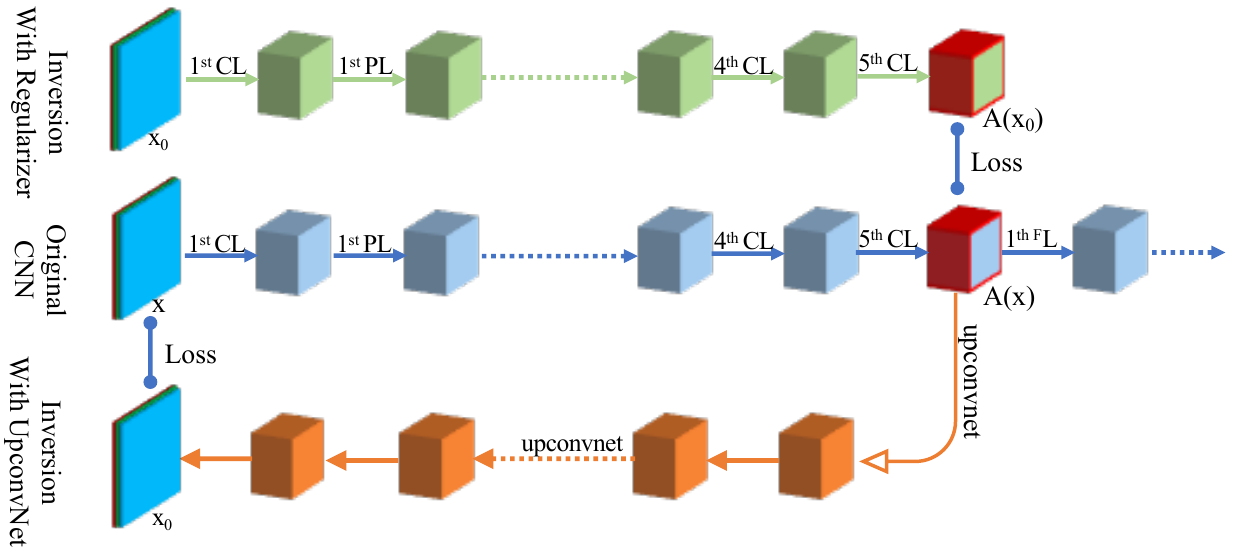}\\
  \caption{The data flow of the two \textit{Network Inversion} algorithms}\label{inver_method}
  \end{center}
\end{figure}

Fig.~\ref{inver_method} shows the network implementation for the two \textit{Network Inversion based Visualization} methods comparing with the original CNN:
	the Regularizer based \textit{Network Inversion} is shown in the upper as denoted in green,
	and the \textit{UpconvNet} based \textit{Network Inversion} is shown in the bottom as denoted in orange,
	while the original CNN is shown in the middle as denoted in blue.

The Regularizer based \textit{Network Inversion} has the same architecture and parameters as the original CNN before the visualization target layer.
	In this case, each pixel of the to be reconstructed image $x_{0}$ is adjusted to minimize the objective loss function error between the target feature map $A(x_{0})$ of $x_{0}$ and the feature map $A(x)$ of the original input image $x$.

For the \textit{UpconvNet} based \textit{Network Inversion}, the \textit{UpconvNet} provides a inverse path for the feature map back to the image dimension.
	The parameters of the \textit{UpconvNet} are adjusted to minimize the objective loss function error between the reconstructed image $x_{0}$ and the original input image $x$.

In the following sections, we will give detailed explanations of the two methods mathematically.

\subsubsection{Regularizer based network inversion}
The fundamental algorithm of Regularizer based \textit{Network Inversion} can be viewed as reconstructing an image $x^{*}$ which minimizes the objective function as following:
\begin{equation}\label{Inversion}
	x^{*} = \mathop{\argmin}_{x} (C\cdot \mathcal{L}({A(x), A(x_{0}))- \lambda (x)}),
\end{equation}
where the loss function $\mathcal{L}$ computes the difference between the two aforementioned feature maps $A(x_{0})$ and $A(x)$.
	The constant $C$ trades off the loss and the regularizer, and the regularizer $\lambda (x)$ restricts the reconstructed image to a natural image.
The loss function is usually defined as a Euclidean distance:
\begin{equation}\label{Euclidean distance}
  \mathcal{L}(A(x), A(x_{0}))= \left \| A(x) - A(x_{0}) \right \| ^{2},
\end{equation}
which is the most commonly used measurement to evaluate the similarity between different images~\cite{Wang:2005:euclidean}.

In order to make the reconstructed images  look closer to the nature images, multiple regularization approaches have been experimentally studied to improve the reconstruction quality, such as $\alpha$-norm, total variation norm (TV) , jittering, and texture or style regularizers~\cite{Rudin:1992:TV,Mordvintsev:2015:Google-dream,Gatys:2015:texture}.
	As an example, for a discrete image data $x \subset R^{H X W}$, the TV norm is given by:
\begin{equation}\label{TV}
 \lambda (x) = \sum_{i,j} ((x_{i,j+1} - x_{i,j})^{2} + (x_{i+1,j} - x_{i,j})^{2})^{\frac{\beta}{2}},
\end{equation}
where the regularizer $\beta$ =1 stands for the standard TV norm that is mostly used in image denoising.
	In this case, the TV norm penalizes the reconstructed images to encourage the spatial smoothness.


Based on such a \textit{Network Inversion} framework, the visualization process can be divided into five steps:

	(1) The visualization target layer's feature maps $A(x)$ of the original input image $x$ and the feature maps $A(x_{0})$ of the to be reconstructed $ x_{0} $ (initialized with noise) are firstly computed.

	(2) The error between the two feature map sets -- $\mathcal{L} (A(x), A(x_{0}))$ is then computed by the objective loss function.

	(3) Guided by the direction of the gradient $\frac{\mathcal{L} (A(x), A(x_{0}))}{ \partial x}$, each pixel of the noise image is changed iteratively to minimize the objective loss function error.

	(4) This process terminates at a specific reconstructed image $x^{*}$, which is used to demonstrate what information is preserve in the visualization target layer.

The Regularizer based \textit{Network Inversion} iteratively tweaks the input noise towards the direction that minimizes the difference between the two feature map sets, while the \textit{UpconvNet} based \textit{Network Inversion} minimizes the image reconstruction error.
In the next section, we will then discuss the \textit{UpconvNet} based \textit{Network Inversion} in detail.

\subsubsection{UpconvNet based network inversion}
Although the Regularizer based \textit{Network Inversion} can reconstruct a image for CNN layer visualization, it still suffer from relatively slow computation speed due to gradient computation.
	To overcome this shortcoming, Dosovitskiy~\textit{et al.} proposed another \textit{Network Inversion} approach, which trained an extra dedicated \textit{Up-convolutional Neural Network} (\textit{UpconvNet}) to reconstruct the image with better image quality and computation efficiency~\cite{Dosovitskiy:2016:CVPR:Network-Inversion}.

The \textit{UpconvNet} can project the low-dimension feature maps back to the image dimension with similar reversed layers as in \textit{DeconvNet}.
As shown in the bottom part of Fig.~\ref{inver_method}, the \textit{UpconvNet} takes the feature maps $A(x)$ as the input, and yields the reconstructed image as the output.

Each layer of the \textit{UpconvNet} are described as follows:

\textit{Reversed Convolutional Layer}: The filters are re-trained in the \textit{UpconvNet} whereas the \textit{DeconvNet} uses the transposed versions of the same convolutional filters.
	Given a training set of images and their feature maps $(x_{i}, A(x_{i}))$, the training procedure can be viewed as:
\begin{equation}\label{Euclidean distance}
  W^{*} =  \mathop{\argmin}_{w} \sum_{i}|| x_{i} - D (A(x_{i}),W)||^{2},
\end{equation}
where the weight $W$ of \textit{UpconvNet} is optimized to minimize the squared Euclidean distance between the input image $x_{i}$ and the output of \textit{UpconvNet} -- $D (A(x_{i}),W)$.

\textit{Reversed Rectification Layer}: The feature maps of \textit{UpconvNet} are also ensured to be positive, with the leaky \textit{relu} nonlinearity of slope 0.2 is applied:
\begin{equation}\label{Euclidean distance}
 A(x) =
  \left\{\begin{matrix}
x    & x \geqslant 0\\
0.2 & x<0
\end{matrix}\right.
 \end{equation}

\textit{Reversed Max-pooling Layer}: The unpooling layers in \textit{UpconvNet} are quite simplified.
	The feature maps are upsampled by a factor of 2, which replaces each value by a 2 $\times$ 2 block with the original value in the top left corner and all other entries equal to zero.

After the training process, we can utilize this \textit{UpconvNet} to reconstruct any input image without computing the gradients.
	Therefore, it dramatically decreases the computational cost and can be applied to various kinds of deep networks.
	In the next section, we will evaluate the visualization results based on these two approaches.

\subsection{Experiments with the network inversion based visualization}
In this section, the experiments of \textit{Network Inversion based Visualization} is demonstrated based on \textit{AlexNet} trained with \textit{ImageNet} dataset. The experiments demonstrate that \textit{Network Inversion based Visualization}  can not only achieve optimal visualization performance, but can be also utilized enhance the CNN design.

\subsubsection{Layer-level visualization analysis}
\begin{figure}[htp]
	\begin{center}
		\includegraphics[width=5in, height=1.3in]{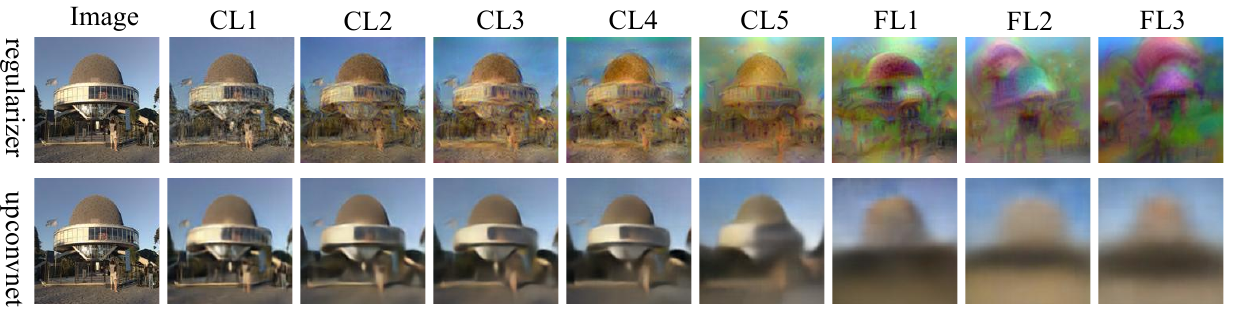}\\
		\caption{\textit{AlexNet} reconstruction by \textit{Network Inversion} with regularizer and \textit{UpconvNet}.
		Adapted from ``Inverting Visual Representations with Convolutional Networks," by A. Dosovitskiy, 2016.}\label{inv_compare}
	\end{center}
\end{figure}

Visualization from layer level can reveal what features are preserved by each layer.
Fig.~\ref{inv_compare}, shows Regularizer and \textit{UpconvNet} based visualization from various layers of \textit{AlexNet}.

From Fig.~\ref{inv_compare}, we can find that:
	1) The visualization from the CLs look similar to the original image, although with increasing fuzziness.
	This indicates that the lower layers preserve much more detailed information, such as colors and locations of objects.
	2) The visualization quality has an obvious drop from the CLs to FLs.
	However, the visualization from higher CLs and even FLs preserve color (\textit{UpconvNet}) and the approximate object location information.
	3) The \textit{UpconvNet} based visualization quality is better than the Regularizer based visualization, especially for the FLs.
	4) The unrelated information is gradually filtered from low layers to high layers.

\subsubsection{Layer level feature map analysis}
\begin{figure}[b]
	\begin{center}
		\includegraphics[width=4in, height=1.8in]{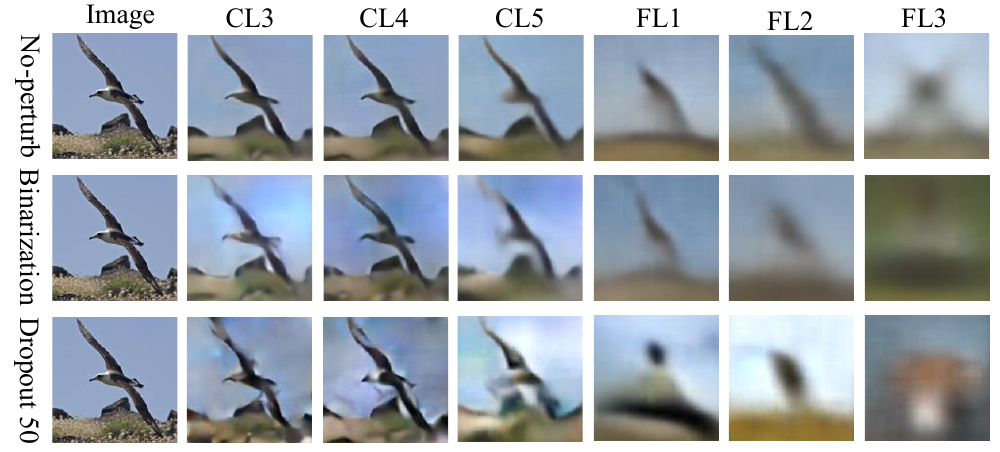}\\
		\caption{\textit{AlexNet} reconstruction by perturbing the feature maps. Adapted from ``Inverting Visual Representations with Convolutional Networks," by A. Dosovitskiy, 2016.}\label{inv4}
	\end{center}
\end{figure}
Based on the layer level analysis, here we further utilize the \textit{Network Inversion based Visualization} to analysis the feature map characteristics.
	In~\cite{Dosovitskiy:2016:CVPR:Network-Inversion}, Dosovitskiy~\textit{et al}. perturbed part of the feature maps in one layer and visualized the reconstructed image from these perturbed features maps.
	The perturbation was performed in two ways:

(1) Binarization: the signs of all feature maps' values are kept, and their absolute values are set to be fixed numbers.
The Euclidean norm of the values are remained unchanged.

(2) Dropout: 50\% of the feature maps' values are set to be zeros and then normalized to keep their Euclidean norm unchanged.

Fig.~\ref{inv4} shows the reconstructed images under the two perturbation approaches in different layers.
	From Fig.~\ref{inv4} we can see that:
	1) In FL1, the binarization hardly changes the reconstruction quality, which means almost all information about the input image is contained in the pattern of non-zero feature maps.
	2) The Dropout changes the reconstructed images a lot.
However, Dosovitskiy~\textit{et al}. also experimentally showed that by Dropouting the 50\% least important feature maps could significantly reduce the reconstruction error, which is even better than not applying any Dropout for most layers.

These observations could be a proof that various CNN compression techniques could achieve optimal performance, such as quantization and filter pruning, due to the considerable amount of redundant information in each layer.
	Hence the \textit{Network Inversion} based Visualization can be used to evaluate the importance of feature maps, and pruning the least important feature maps for network compression.

\subsection{The summary}
The \textit{Network Inversion based Visualization}  projects a specific layer's feature maps back to the image dimension, which provides insights into what features a specific layer would preserve.
	Additionally, by perturbing some feature maps for visualization, we can verify the CNN preserved a lot redundant information in each layer, and therefore further optimize the CNN design.

\section{Visualization by network dissection}
\textit{\vspace{1mm}\\\-\hspace{0.5cm}Evaluate the correlation between each convolutional neuron or multiple neurons with a specific semantic concept.}

\subsection{The overview}

In previous sections, multiple visualization methods were demonstrated to reveal the visual perceptible patterns that a single neuron or layer could capture.
	However, there is still a missing link between the visual perceptible patterns and the clear interpretable semantic concepts.
\begin{figure}[b]
	\begin{center}
		\includegraphics[width=4.5in, height=1.6in]{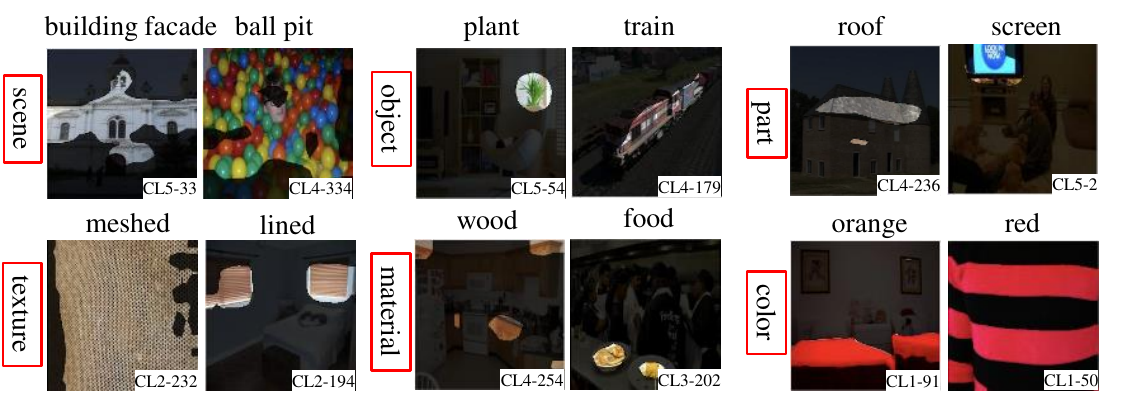}\\
		\caption{The \textit{Broden} images that activate certain neurons in \textit{AlexNet}}
		\label{Broden}
	\end{center}
\end{figure}

Hence, Bau \textit{et al.} proposed the \textit{Network Dissection}, which directly associates each convolutional neuron with a specific semantic concept, such as color, textures, materials, parts, objects, and scenes.
	The correlation between the neuron and the semantic concept is measured by seeking for the neuron that strongly responses to particular image content with specific semantic concepts.
	A heterogeneous image dataset --- \textit{Borden}, provides the images with specific semantic concepts labeled corresponding to local content.
	A set of \textit{Broden} examples are shown in Fig.~\ref{Broden}, in which the semantic concepts are divided into six categories highlighted with red boxes.
	Each semantic category may cover various classes, for example, the object category contains plant, train, \textit{etc}.
	At the lower right corner of each example in Fig.~\ref{Broden}, the semantic corresponding neuron is also identified.
	We can also see that black masks are introduced to cover the image content that is not related to the assigned semantics.
	Here, it is the proposed \textit{Network Dissection} that generates these black masks.

The development of the \textit{Network Inversion} progressively connects the semantic concepts to different component levels in a CNN.
	The fundamental algorithm of \textit{Network Inversion} illustrated the correlation between one semantic concept and one individual neurons.
	Such a correlation was based on an assumption that each semantic concept can be assigned to a single neuron~\cite{Gonzalez:2018:semantic}.
	Later, further \textit{Network Inversion} works revealed that the feature representation can be distributed, which indicated that one semantic concept could be represented by multiple neurons' combination~\cite{Agrawal:2014:multi_unit,Zhou:2014:ICLR:Object-Detectors}.
	Hence, following~\cite{Bau:2017:CVPR:Network-Dissection}'s paradigm, Fong~\textit{et al.} proposed another \textit{Network Inversion} approach, namely \textit{Net2Vec}, which visualized the semantic concepts based on neuron combinations~\cite{Fong:2018:net2vec}.

Both methods provide comprehensive visualization results on interpreting CNN hidden neurons.

\subsection{The algorithm}
In this section, we introduce two \textit{Network Dissection} methods, one method assigns the semantic concept to each individual neuron, while the other builds the correlation between the neuron combinations and the semantic concepts.

\subsubsection {Network dissection for the individual neuron}
\begin{figure}[b]
	\begin{center}
		\includegraphics[width=5in, height=2in]{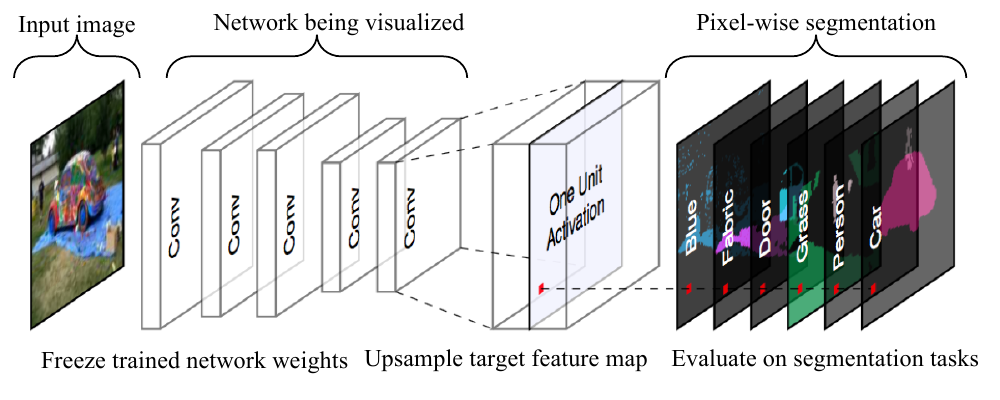}\\
		\caption{Illustration of network dissection for measuring semantic alignment of neuron in a given CNN. Adapted from ``Network Dissection: Quantifying Interpretability of Deep Visual Representations," by D. Bau, 2017.}\label{dissection_schem}
	\end{center}
\end{figure}

The algorithm of \textit{Network Dissection} for the individual neuron evaluates the correlation between each single neuron and the semantic concept.
	Specifically, every individual neuron is evaluated as a segmentation task to every semantic concept.

The evaluating process is shown in Fig.~\ref{dissection_schem}.
	The input image fetched from the \textit{Broden} dataset contains pixel-wise annotations for the semantic concepts, which provides the ground truth segmentation masks.
	The target neuron's feature map is upsampled to the resolution of the ground truth segmentation masks.
	Then the \textit{Network Dissection} works by measuring the alignment between the upsampled neuron activation map and the ground truth segmentation masks.
	If the measurement result is larger than a threshold, the neuron can be viewed as a visual detector for specific semantic concept.

This process can be described as follows:

	(1) The feature map $A_{f}(x)$ of every neuron $f$ is computed by feeding in every input image $x$ from \textit{Broden}.
	So, the distributions of the activation scores $p(a_{k})$ of neuron activation over all images in \textit{Broden} are computed.

	(2) The top activation maps from all feature maps are selected as valid map regions corresponding to neuron's semantics by setting an activation threshold that $P(a_{f} > T_{f}) = 0.005$.

	(3) To match the low-resolution valid map to the ground truth segmentation mask $L_{c}$ for some semantic concept $c$, the valid map is upsampled:
\begin{equation}\label{mask}
	M_{f}(x) = S(A_{f}(x) > T_{f}),
\end{equation}
where $S$ denotes a bilinear interpolation function.

	(4) The accuracy of neuron $f$ in detecting semantic concept $c$ is determined by the intersection-over-union (IoU) score:
\begin{equation}\label{IoU}
IoU_{f,c} = \frac{\sum |M_{f}(x) \bigcap L_{c}(x)|}{\sum |M_{f}(x)\bigcup L_{c}(x)| },
\end{equation}
where $L_{c}(x)$ denotes the ground-truth mask of the semantic concept $c$ on the image $x$.
	If $IoU_{f,c}$ is larger than a threshold (0.04), we consider the neuron $f$ as a visual detector for concept $c$.
	The $IoU$ score indicates the accuracy of neuron $f$ in detecting concept $c$.
	Finally, every neuron's corresponding semantic concept can be determined by calculating its $IoU$ score.

Hence, the \textit{Network Dissection} for the individual neuron can automatically assign a semantic concept to each convolutional neuron.

\subsubsection {Network dissection for the neuron combinations}
Instead of interpreting the individual neuron, Fong~\textit{et al}.~\cite{Fong:2018:net2vec} proposed~\textit{Net2Vec} to evaluate the correlation between the neuron combinations and the semantic concepts.
	They implemented this approach as a segmentation task by using convolutional neuron combinations.
	Specifically, a learnable concept weight $w$ is used to linearly combine the threshold based activation.
	And, it is passed through the sigmoid function $\sigma (x) = 1/(1 + exp( - x)) $ to predict a segmentation mask $M (x; w)$:
\begin{equation}\label{neuron combinations}
	M (x;w) = \sigma (\sum_{k} w_{k} \cdot \mathbb{I}(A_{f}(x) > T_{f})),
\end{equation}
where $k$ is the number of neurons in a layer, and $\mathbb{I} (\cdot)$ is the indicator function.
	This function selects a subset of neurons in one layer whose activation is larger than the threshold.
	Hence, this subset of neurons can be used to generate the activation mask for specific semantic.

Fong~\textit{et al} experimentally found that, for the segmentation task, materials and parts reached near optimal performance around $k=8$, which was much more quickly than that of objects $k=16$.
	For each concept $c$, the weights $w$ were learned using stochastic gradient descent with momentum to minimize per-pixel binary cross entropy loss.

Similar to the single neurons case, the $IoU_{com}$ score for neuron combinations is computed as well:
\begin{equation}\label{IoU}
	IoU_{com} = \frac{\sum |M_{f,w}(x) \bigcap L_{c}(x)|}{\sum |M_{f,w}(x)\bigcup L_{c}(x)|}.
\end{equation}
	If the $IoU_{com}$ score is larger than a threshold, we consider this neuron combinations as a visual detector for concept $c$.

In fact, using the learned weights to combine neurons outperforms using a single neuron on the segmentation tasks.
	As a result, the generated segmentation masks demonstrate more complete and obvious objects in the image.
	In the next section, we demonstrate the visualization results by these two approaches.

\subsection{Experiments with the network dissection based visualization}
The ~\textit{Network Dissection} can be applied to any CNN using a forward pass without the need for training or computing the gradients.
In this section, we demonstrate the \textit{Network Dissection based Visualization} results based on \textit{AlexNet} trained with \textit{ImageNet}.

\subsubsection{Network dissection for the individual neuron}
\begin{figure}[tp]
	\begin{center}
		\includegraphics[width=5in, height=2.5in]{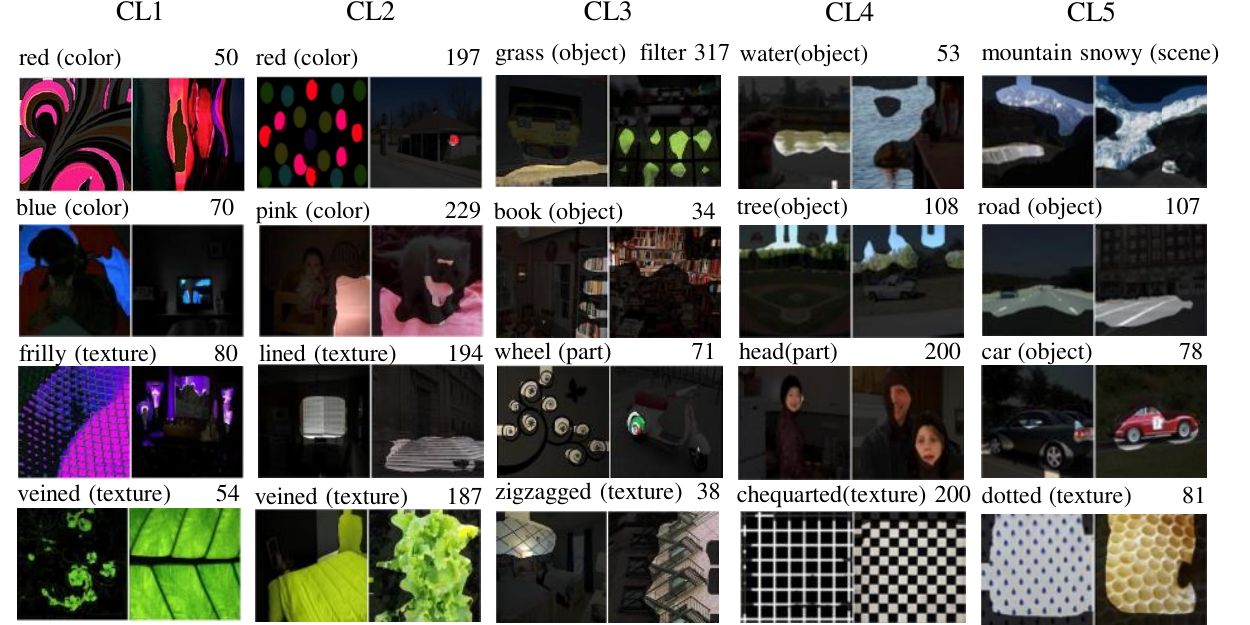}
		\caption{\textit{AlexNet} visualization by \textit{Network Dissection}}\label{dissection_single}
	\end{center}
\end{figure}
\begin{figure}[htp]
	\begin{center}
		\includegraphics[width=5in, height=2in]{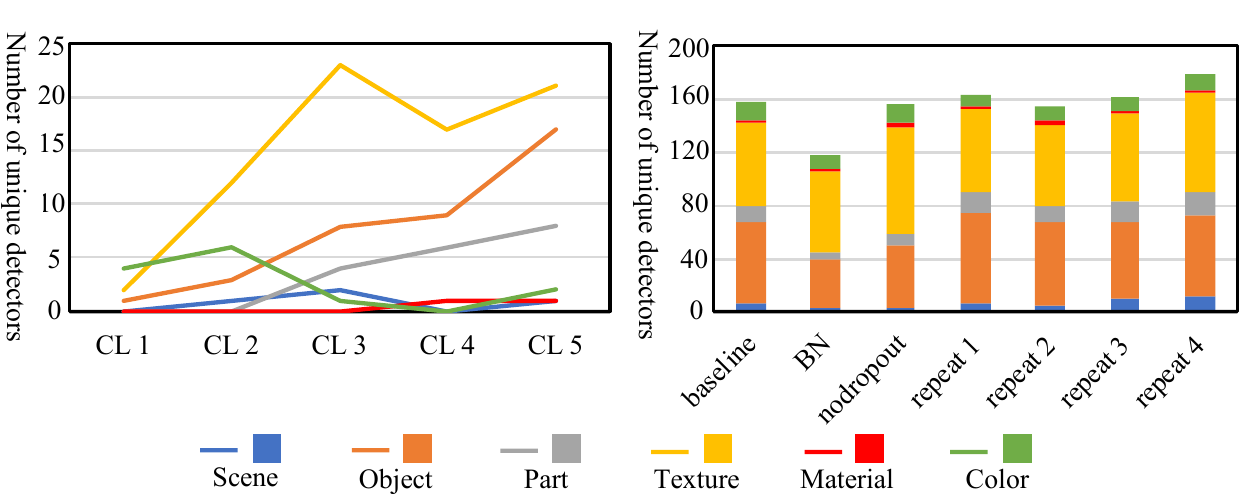}\\
		\caption{Semantic concept emerging in each layers and under different training conditions}\label{dissection_chart}
	\end{center}
\end{figure}

The visualization results of the individual neuron is demonstrated in the Fig.~\ref{dissection_single}.
	In each column, four individual neurons along with two \textit{Broden} images are shown in each CL.
	For each neuron, the top left shows the predicted semantic concepts, and the top right shows the neuron number.
	As mentioned, if the $IoU_{com}$ score is larger than a threshold, we consider this neuron as a visual detector for concept $c$.
	Each layer's visual detector number is summarized in the left part of Fig.~\ref{dissection_chart}, which counts the number of unique concepts matched with neurons.

From the figures, we can find that:
	1) Every image highlights the regions that cause the high neural activation from a real image.
	2) The predicted labels match the highlighted regions pretty well.
	3) From the number of the detector summary, the color concept dominates at lower layers (CL 1 and CL 2), while more object and texture detectors emerge in CL 5.

Compared with previous visualization methods, we conclude that the CNNs could detect the basic information, such as color and texture by all layer neurons rather than lower layer neurons.
	And the color information can be preserved even in higher layers, since many color detectors are also found in these layers.

\subsubsection{Interpretability under different training conditions}
\begin{figure}[hb]
	\begin{center}s
		\includegraphics[width=5in, height=2.8in]{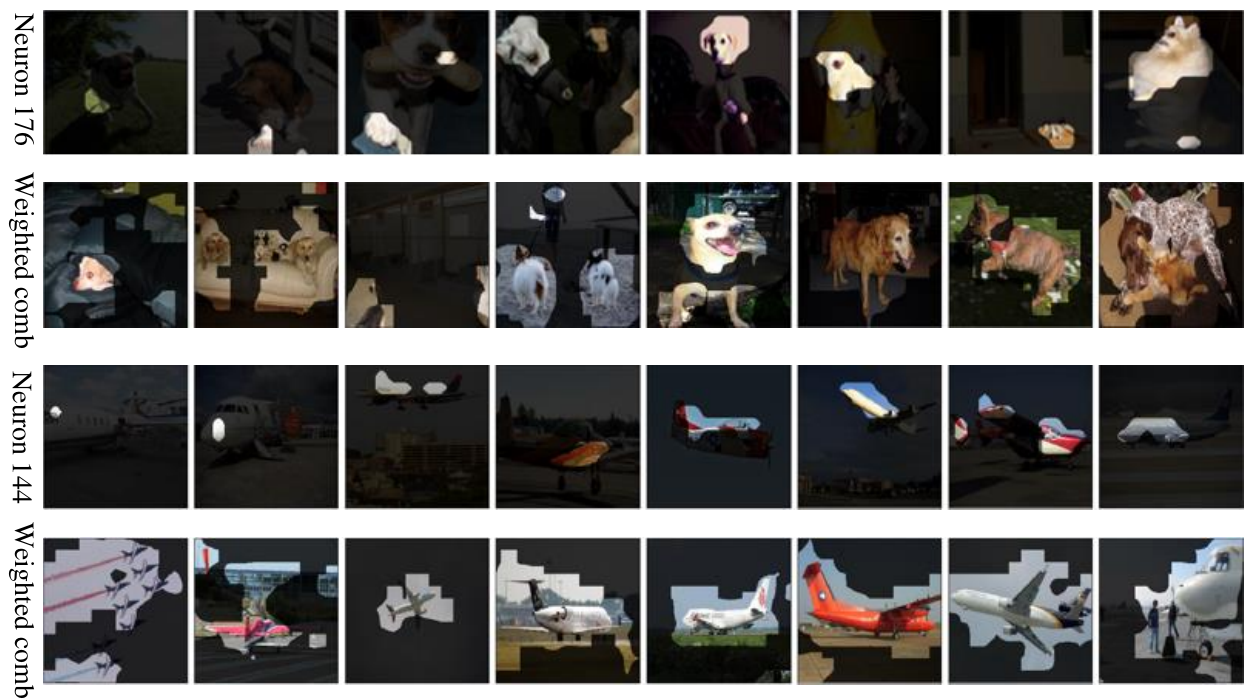}\\
		\caption{\textit{Network Dissection} with single neuron and neuron combinations. Adapted from ``Net2Vec: Quantifying and Explaining how Concepts are Encoded by Filters in Deep Neural Networks," by R. Fong, 2018.}\label{dissection_comb}
	\end{center}
\end{figure}

The training conditions, such as the number of training iterations
could also affect the representation learning of the CNNs.
	In~\cite{Bau:2017:CVPR:Network-Dissection}, Bau~\textit{et al.} evaluated the effects on the interpretability of various state-of-the-art CNNs by using different training conditions, such as Dropout, batch normalization~\cite{Ioffe:2015:batch_norm} and random initialization.
	As shown in the right part of Fig.~\ref{dissection_chart}, the NoDropout indicates the Dropout in the FC layers of the baseline model -- \textit{AlexNet} is removed.
	The BN indicates the batch normalization are applied at each CL.
	While, repeat1, repeat2 and repeat3 indicate randomly initialize the weights with the number of training iterations.

From Fig.~\ref{dissection_chart}, we can observe that:
	1) The network shows similar interpretability under different initialization configurations.
	2) For the network without Dropout applied, more texture detectors emerge, but fewer object detectors.
	3) The batch normalization seems to decrease interpretability significantly.
	Overall, the Dropout and batch normalization can improve the classification accuracy.
	From the visualization perspective, the network tend to capture basic information without dropout.
	And, the batch normalization potentially decrease the feature diversity.

With such an evaluation, we can find that the \textit{Network Dissection based Visualization} could effectively applied into evaluating different CNN optimization methods with a perspective of network interpretability.

\subsubsection{Network dissection for the neuron combinations}
The visualization results by using combined neurons are shown in Fig.~\ref{dissection_comb}.
	The first and third rows are the segmentation results by the individual neuron, while the second and fourth rows are segmented by neuron combinations.
	As we can see, for semantic visualization of ``dog" and ``airplane" using the weighted combination method, the predicted masks are informative and salient for most of the examples.
	This suggests that, although neurons that are specific to a concept can be found, these do not optimally represented or fully cover with the concept.
\subsection{The summary}
The \textit{Network Dissection} is a distinguished visualization method to interpret the CNNs, which can automatically assign semantic concepts to internal neurons.
	By measuring the alignment between the unsampled neuron activation and the ground truth images with semantic labels, \textit{Network Dissection} can visualize the types of semantic concepts represented by each convolutional neuron.
	The~\textit{Net2Vec} also verifies that the CNNs feature representation is distributed.
	Additionally, the \textit{Network Dissection} can be utilized to evaluate various training conditions, which shows the training conditions can have a significant effect on the interpretability of the representation learned by hidden neurons.
	Hence, it is another representative example for CNN visualization and CNN optimization.

\section{CNN visualization application}
In this section, we review some practical applications of the CNN visualization.
In fact, due to its ability to interpret the CNNs, CNN visualization has became an effective tools to reveal the differences between the way CNNs and humans recognize objects.
We also applied the ~\textit{Network Inversion} into an art generation algorithm called style transfer.

\subsection{Visualization analysis for CNN adversarial noises}
\begin{figure}[b]
	\begin{center}
		\includegraphics[width=3.7in, height=1.3in]{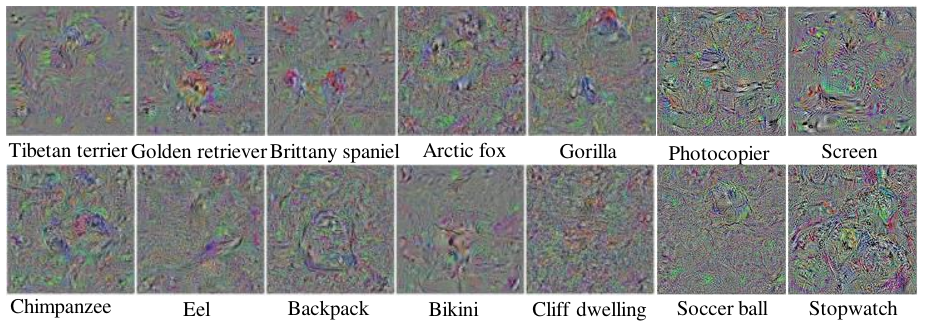}\\
		\caption{Adversarial noises that manipulate the CNN classification}\label{fool}
	\end{center}
\end{figure}
The CNNs has\break achieved impressive performance on a variety of computer vision related tasks.
	The CNNs are able to classify objects in images with even beyond human-level accuracy.
	However, we can still produce images with adversarial noises to attack the CNN for classification result manipulation, while the noises are  completely unperceivable by the human vision recognition.
	Such adversarial noises could manipulate the results of the state-of-the-art CNNs with a successful rate of 99.99\%~\cite{Nguyen:2015:fool}.
	Hence, questions naturally arise as what differences remain between the CNNs and the human vision.

The \textit{Activation Maximization} can be well utilized to examine those adversarial noises.
	As shown in the Fig.~\ref{fool}, the adversarial noises are generated by directly maximizing the final layer output for classes via gradient ascent. And continues until the CNNs confidence for the target class reaches 99.99\%.
	Adding regularization makes images more recognizable, still far away from human interpretable images, but results have slightly lower confidence scores.

CNNs recognize these adversarial noises as near-perfect examples of recognizable images, which indicates that the differences between the way CNNs and humans recognize objects.
	Although CNNs are now being used for a variety of machine learning tasks.
	It is required to understand the generalization capabilities of CNNs, and find potential ways to make them robust.
	And the visualization is an optimal way to directly interpret those adversarial threat potentials.

\subsection{Visualization based adversarial examples}
\begin{figure}[b]
	\begin{center}
		\includegraphics[width=3in, height=1.4in]{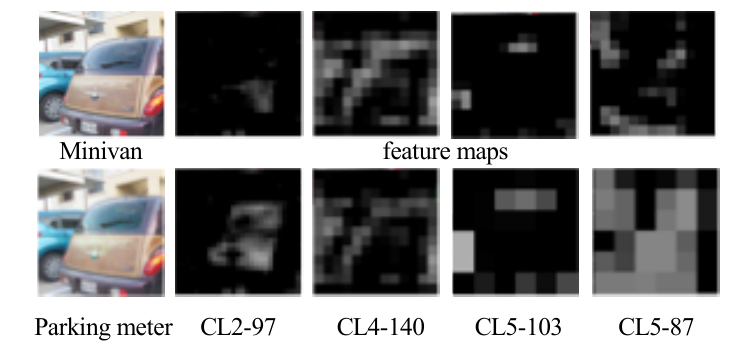}
		\vspace{-4mm}
		\caption{Adversarial example visualization}
		\label{adversarial}
	\end{center}
\end{figure}
In this section, we present another analysis approaching to demonstrate how the CNNs and the human vision differ.
	Recent studies show that the adversarial noises can be well embedded into images forming adversarial examples, which can manipulate the CNN classification results without noticing.
	Some adversarial examples are shown in  Fig.~\ref{adversarial}.

To improve the robustness of the CNNs, the traditional techniques such as batch normalization and Dropout, generally do not provide a practical defense against adversarial examples.
	Some other strategies such as adversarial training and defensive distillation has been proposed to defense against the adversarial examples, which achieve state-of-art results.
	Yet even these specialized algorithms can easily be broken by giving more delicate adversarial examples.

The visualization can provide a solution to further uncover the mystery of adversarial examples.
	The activation maps of four convolutional neurons are shown in the Fig.~\ref{adversarial}.
	We can observe that the visualized feature maps have been changed a lot by the adversarial noises, especially for the CL2-97 and CL5-87.
	Hence, even the human vision can hard perceive the adversarial examples, the visualization could provide a significantly effective detection approach.
	The visualization analysis of the adversarial examples also reveal another major difference between CNN and the human vision: the imperceptible patterns can be captured by the CNNs and greatly affect the classification results.

\subsection{Visualization analysis for style transfer}
As we discussed in the Section 5, we can visualize the information each layer preserved from the input image, by reconstructing the image from feature maps in one layer.
	And we know the higher layers capture the high-level content in terms of objects and their arrangement in the input image but do not constrain the exact pixel values of the input image.
	These feature maps in higher layers can be referred as the \textit{style} of an image~\cite{Gatys:2015:neural_style}.
\begin{figure}[t]
	\begin{center}
		\includegraphics[width=3.8in]{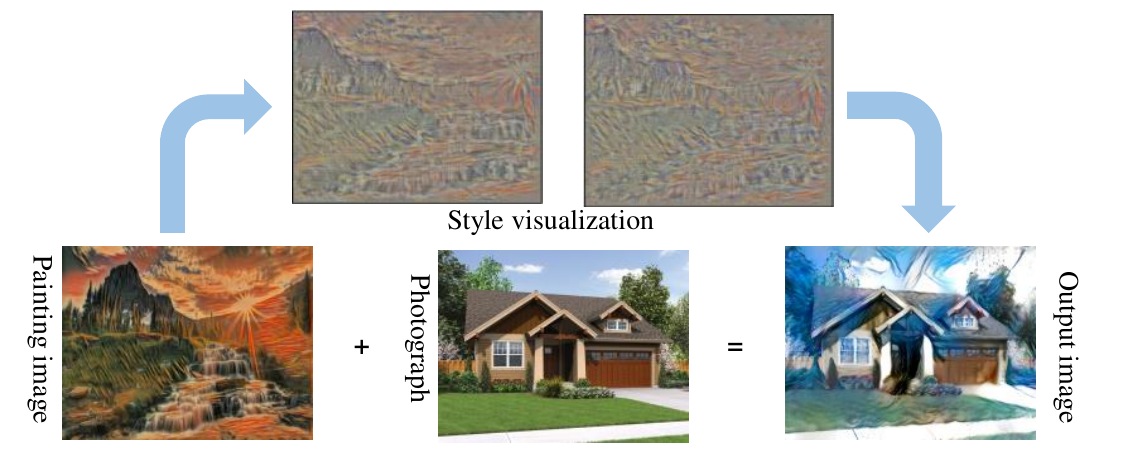}
		\vspace{-2mm}
		\caption{Style transfer example}
		\label{Style}
	\end{center}
\end{figure}

As shown in Fig.~\ref{Style}, the style transfer generates the output image that combines the \textit{style} of a painting image $a$ with the content of a photograph $p$.
We can utilize the ~\textit{Network Inversion} visualize the style of the painting image.
The process can be described viewed as jointly minimizing:
\begin{equation}\label{Inversion}
	\mathcal{L}_{total} = \alpha \mathcal{L}_{content}({A(p), A(x)}) + \beta \mathcal{L}_{style}({A(a), A(x)}),
\end{equation}
where the $A(p)$ is the feature maps of the photograph in one layer and $A(a)$ is the feature maps painting image in multiple layers.
The ratio $\frac{\alpha}{\beta}$ of content loss and style loss adjust the emphasis on matching the content of the photograph or the style of the painting image.

This process renders the photograph in the style of the painting image, which the appearance of the output image resembles the style of painting image, and the output image shows the same content as the photograph.

\subsection{Summary}
In this section, we briefed several applications of the CNN visualization beyond the scope of CNN interpretability enhancement and optimization. While more visualization applications still remained undiscussed.
	We do believe that the visualization could contribute to the CNN analysis in more and more perspectives.
\section{Conclusion}
In this paper, we have reviewed the latest developments of the CNN visualization methods.
	Four representative visualization methods are delicately presented, in terms of structure, algorithm, operation, and experiment, to cover the state-of-the-art research results of CNN interpretation.

Trough the study of the representative visualization methods, we can tell that:
	The CNNs do have hierarchical feature representation mechanism that imitates the hierarchical organization of the human visual cortex.
	Also, to reveal the CNN interpretation, the visualization works need to take various perspectives regarding different CNN components.
	Moreover, the better interpretability of the CNN introduced by visualization could practically contribute to the CNN optimization.

Hence, as the CNNs continually dominate the computer vision related tasks, the CNN visualization would play a more and more important role for better understanding and utilizing the CNNs.
\section*{Acknowledgments}
This work was supported in part by NSF CNS-1717775.


\providecommand{\href}[2]{#2}
\providecommand{\arxiv}[1]{\href{http://arxiv.org/abs/#1}{arXiv:#1}}
\providecommand{\url}[1]{\texttt{#1}}
\providecommand{\urlprefix}{URL }

\medskip

Received  October 2017; revised December 2017.
\medskip

\end{document}